\documentclass{article} 

\usepackage{fullpage}
\usepackage{amsmath,amssymb, amsthm}
\usepackage[hyphens]{url}  

\usepackage[numbers]{natbib}  
\usepackage{caption} 

\usepackage{floatflt}
\usepackage{wrapfig}
\usepackage{subcaption}
\usepackage{comment}

\usepackage{hyperref}

\usepackage{algorithm}
\usepackage{algorithmic}
\usepackage{bm}
\usepackage[pdftex]{graphicx}
\newtheorem{definition}{Definition}
\newtheorem{theorem}{Theorem}

\usepackage{xcolor}
\usepackage{booktabs}

\DeclareMathOperator*{\argmin}{arg\,min}

\title{Differentially Private Normalizing Flows for \\ Privacy-Preserving Density Estimation}
\author {Chris Waites\footnotemark[1] \and Rachel Cummings\footnotemark[2]}

\begin{document}

\maketitle

\renewcommand{\thefootnote}{\fnsymbol{footnote}}
\footnotetext[1]{Department of Computer Science, Stanford University. Email: \texttt{waites@stanford.edu}}
\footnotetext[2]{Department of Industrial Engineering and Operations Research, Columbia University. Email: \texttt{rac2239@columbia.edu} Supported in part by NSF grants CNS-1850187 and CNS-1942772 (CAREER), a Mozilla Research Grant, and a JPMorgan Chase Faculty Research Award. This work was completed while R.C. was at Georgia Institute of Technology.}

\renewcommand{\thefootnote}{\arabic{footnote}}


\begin{abstract}
Normalizing flow models have risen as a popular solution to the problem of density estimation, enabling high-quality synthetic data generation as well as exact probability density evaluation. However, in contexts where individuals are directly associated with the training data, releasing such a model raises privacy concerns. In this work, we propose the use of normalizing flow models that provide explicit differential privacy guarantees as a novel approach to the problem of privacy-preserving density estimation. We evaluate the efficacy of our approach empirically using benchmark datasets, and we demonstrate that our method substantially outperforms previous state-of-the-art approaches.  We additionally show how our algorithm can be applied to the task of differentially private anomaly detection.
\end{abstract}

\section{Introduction}\label{s.intro}


The task of density estimation requires constructing an estimate of an unknown probability density function, given observed data. This density estimate can then be used to perform a variety of relevant analysis tasks, including log likelihood evaluation and synthetic data generation. In settings involving sensitive data, the construction and subsequent release of such an estimate could potentially leak private information. Without a rigorous privacy guarantee, nothing prevents a model from memorizing a row in the training set, assigning disproportionate density to a point, or any other vulnerability due to arbitrary analysis of the learned parameters. Since density estimation remains a task of interest to the modeling community, continued attention is required to develop privacy-preserving methods for density estimation.

Differential privacy \citep{DMNS06} has emerged as the predominant privacy notion in the context of statistical data analysis. At a high level, differentially private analyses limit the extent to which the distribution of outputs can change due to the inclusion or exclusion of any one individual from the analysis. Algorithms which adhere to this notion exhibit a number of desirable properties, including privacy guarantees which hold regardless of the auxiliary information an adversary may have and composition of privacy guarantees across multiple analyses. Hence differential privacy acts as a compelling gold standard in the design of privacy-preserving analyses.

Tools for density estimation have held longstanding interest due to their versatility. Their ability to address a wide range of distributional learning tasks is precisely why the existence of an accurate and privacy-preserving density estimation is surprising. For example, privately constructing such a model implicitly yields a differentially private approach to anomaly detection---a task of substantial investigation \citep{inbook, okada2015differentially, 6754007}---as an immediate application of likelihood inference. In addition, given that density estimators often enable efficient sampling, such a model would yield a method for privacy-preserving synthetic data generation. This task in particular has been of longstanding interest to the privacy community \citep{Surendra2017ARO} as it addresses many of the limitations imposed by the query-release model \citep{dwork2008differential} by allowing large numbers of arbitrary analyses. Privately generating a synthetic dataset only incurs a fixed privacy cost during the training process; all subsequent queries on the synthetic data are automatically differentially private due to the privacy notion's post-processing guarantee, so the privacy cost does not scale with the number of downstream analyses performed.

Normalizing flow models are an attractive approach to the task of density estimation due to their empirical ability to approximate arbitrary, high-dimensional distributions. These models approach the task of density estimation via a transformation on a chosen base density by a sequence of invertible, non-linear transformations, enabling density querying on the resulting distribution via an application of the change-of-variables formula. Approaches to density estimation in this manner include: Non-linear Independent Components Estimation (NICE) \citep{Dinh2014NICENI}, Real NVP \citep{dinh2016density}, Glow \citep{kingma2018glow}, and Masked Autoregressive Flows (MAF) \citep{NIPS2017_6828}. Until this work, it was an open question whether normalizing flow models could be constructed in a differentially private manner to handle the task of privacy-preserving density estimation, combining the rigorous guarantees of differential privacy with the strong empirical performance exhibited by normalizing flows. 

In this work we propose the use of normalizing flow models trained in a differentially private manner as a novel approach to the task of privacy-preserving density estimation. We provide an algorithm (DP-NF, Algorithm \ref{alg:DP-NF} in Section \ref{s.algo}) that privately optimizes the model parameters via gradient descent using DP-SGD \citep{Abadi_2016}, which adds Gaussian noise to clipped gradient updates ensure differential privacy. Additionally, we achieve tighter privacy guarantees than established in previous work \citep{Abadi_2016} via composition with the recently introduced notion of Gaussian differential privacy \citep{DBLP:journals/corr/abs-1905-02383}. We apply this optimization to the parameters of a Masked Autoregressive Flow \citep{NIPS2017_6828}, our primary architecture of consideration, and achieve empirical results (Section \ref{s.exp}) which convincingly outperform previous approaches. Further, we show that our algorithm can be applied to solve the problem of differentially private anomaly detection (Section \ref{s.anomaly}), and show that it leads to better true/false positive rates than existing private methods. 


\subsection{Related Work}\label{s.rel}

Gaussian mixture models (GMMs) are known to be a particularly strong density estimation tool \citep{papamakarios2019neural} since they are a \emph{universal approximator of densities} --- that is, they are able to approximate any density function arbitrarily well given a sufficient number of components \citep{mclachlantext}. They approach the task of density estimation by modeling the data distribution as a weighted sum of Gaussian distributions. The first differentially private algorithm for learning the parameters of a Gaussian mixture model comes from the work of \citep{10.1145/1250790.1250803}, which uses their \emph{sample-and-aggregate} framework to convert non-private algorithms into private algorithms, applied to the task of learning mixtures of Gaussians. However, their approach exhibits strong assumptions on the range of the parameter space and assumes a uniform mixture of spherical Gaussians. Follow-up work of \citep{dpgmmkamath} proposes a modernized approach which improves upon the sample complexity of the aforementioned work and removes the strong a priori bounds on the parameters of the mixture components, although it makes the assumption that the components of the mixture are well-separated. 

There has also been work in learning the parameters of a Gaussian mixture model through differentially private variants of the expectation maximization (EM) algorithm. One notable instance of this is DPGMM \citep{7590445}, which achieves a privacy guarantee at each iteration of EM through the application of calibrated Laplace noise to the estimated model parameters following each maximization step. These individual privacy guarantees are then combined into an overall privacy guarantee via sequential composition, i.e., by taking the sum of privacy parameters in each iteration. The work of \citep{pmlr-v54-park17c} introduces DP-EM, a general framework for privacy-preserving optimization via expectation maximization. Their approach follows a conceptually similar idea of applying either calibrated Laplace or Gaussian noise to the model parameters at the end of each EM iteration. They apply this method to learning mixtures of Gaussians, henceforth referred to as DP-MoG, and they demonstrate significantly better privacy guarantees through composition via the moments accountant and zero-concentrated differential privacy (zCDP) \citep{DBLP:journals/corr/BunS16}. Given that their work makes no notable assumptions about the task and provides an empirical evaluation of their method, this is the most comparable approach to our own. As such, it is used as a baseline in our experimental results.

In addition, we take note of more classical approaches to the task of privacy-preserving density estimation. One of the simplest yet most widely used methods for density estimation is through the use of histograms, and previous work \citep{chawla2005histograms, 6228070} has investigated their private estimation. Unfortunately, such an approach scales poorly with the dimension and complexity of the distribution while asserting an unrealistic discretization of the space. Kernel density estimation is another closely related approach, often characterized as the smooth analog to the classical discrete histogram. The work of \citep{10.5555/2567709.2502603} proposes a method for privately querying the density of such an estimator through the addition of calibrated Gaussian noise. As a non-parametric approach, it has the drawback that it requires storage of the entire dataset at test time to enable querying (proving impractical for large-scale datasets) while still degrading similarly with dimension.

There have also been a number of deep learning based approaches to generative modeling which vary in their relevance. Although work of this nature technically allows for both sampling and likelihood evaluation, it does not allow for \textit{exact} likelihood inference as is the case for mixtures of Gaussians and normalizing flows. There is also expansive literature concerning differentially private approaches to training Generative Adversarial Networks, yet these methods are strictly limited to sampling and do not provide a straightforward approach to likelihood inference.

Finally, we include a brief overview of the extensive literature concerning density estimation via normalizing flows. One important subset are those characterized by \emph{coupling layers}: transformations which partition the dimensions of its input and map them in a way that retains invertibility and a tractable Jacobian. This includes Non-linear Independent Components Estimation (NICE) \citep{Dinh2014NICENI}, as well as its subsequent generalization Real NVP \citep{dinh2016density}. Another notable approach, Glow \citep{kingma2018glow}, makes use of such coupling layers while also proposing the use of an invertible weight matrix decomposition to generalize the notion of permutation layers. Alternatively, some make use of \emph{autoregressive transformations}, which are transformations that utilize the chain rule of probability to represent a joint distribution as a product of its conditionals. Such models include Masked Autoregressive Flow (MAF) \citep{NIPS2017_6828}, a generalization of Real NVP optimized for density estimation, as well as its closely related Inverse Autoregressive Flow \citep{DBLP:journals/corr/KingmaSW16} optimized for variational inference, among others \citep{oliva2018transformation, DBLP:journals/corr/abs-1804-00779,fakoor2020trade}.

\section{Preliminaries}\label{s.prelim}

\subsection{Normalizing Flows}\label{s.normflow}

Let $p(\cdot)$ be the probability density function characterizing an unobservable distribution of interest, and let $\bm{X} = \{\bm{x}^{(1)}, \ldots, \bm{x}^{(n)}\}$ be $n$ observed i.i.d. samples from this distribution. The task of density estimation is to find an approximation of $p(\cdot)$ via some model $p_{\bm{\theta}}(\cdot)$ given $\bm{X}$. In the context of normalizing flows, this model is characterized by a prior distribution $q(\cdot)$, chosen to exhibit a simple and tractable density (e.g., the spherical multivariate Gaussian distribution), and a sequence of $K$ bijective functions $f_{\bm{\theta}} = f_{1} \circ f_{2} \circ \ldots \circ f_{K}$, parameterized fullyby $\bm{\theta}$. The function $f_{\bm{\theta}}$ acts as a transformation between the prior distribution $q(\cdot)$ and the approximated distribution $p_{\bm{\theta}}(\cdot)$.

Given such a model, it can be used to efficiently sample $\bm{x}\sim p_{\bm{\theta}}$ by first sampling $\bm{z} \sim q$ and then transforming the sample as $\bm{x} = f_{\bm{\theta}}(\bm{z})$.  If $p_{\bm{\theta}}$ is a good approximation of $p$, then this generative process gives an efficient (approximate) oracle for sampling from the unknown distribution.

Since $f_{\bm{\theta}}$ is invertible, one can also perform exact likelihood evaluation on observed points from the data distribution via the change of variables formula, as follows: 
\begin{align*}
\log p_{\bm{\theta}}(\bm{x})  &= \log q(f_{\bm{\theta}}^{-1}(\bm{x})) + \log \left| \det \left( \frac{\partial f_{\bm{\theta}}^{-1}(\bm{x})}{\partial \bm{x}} \right) \right|\\ &= \log q(f_{\bm{\theta}}^{-1}(\bm{x})) + \sum_{i = 1}^{K} \log \left| \det \left( \frac{\partial f_{i}^{-1}(\bm{x})}{\partial \bm{x}} \right) \right|.
\end{align*}

Finding a good approximation $p_{\bm{\theta}}$ is achieved through optimization of $\bm{\theta}$ to minimize the negative log likelihood of the observed dataset: 
\begin{equation}\label{eq.loss}
\mathcal{L}(\bm{\theta}) = -\frac{1}{n} \sum_{i=1}^{n} \log p_{\bm{\theta}}(\bm{x}^{(i)}).
\end{equation} 
In practice, one will typically find the MLE $\bm{\theta}^* = \argmin_{\bm{\theta}} \mathcal{L}(\bm{\theta})$ using some non-convex optimization method, such as stochastic gradient descent.

\subsection{Differential Privacy}\label{s.dp}

Differential privacy \citep{DMNS06} has become the gold standard for ensuring the privacy of statistical analyses applied to sensitive databases. At a high level, it ensures that changing a single entry in the database will have only a small effect on the distribution of analysis results.

\begin{definition}[\citep{DMNS06}]\label{def.dp}
A randomized algorithm $\mathcal{M} : \mathcal{D} \rightarrow \mathcal{R}$ satisfies $(\varepsilon, \delta)$-differential privacy (DP) if for any two input database $D, D' \in \mathcal{D}$ that differ in a single entry and for any subset of outputs $\mathcal{S} \subseteq \mathcal{R}$, it satisfies, 
\[ \Pr[\mathcal{M}(D) \in \mathcal{S}] \leq e^\varepsilon \Pr[\mathcal{M}(D') \in \mathcal{S}] + \delta. \]
\end{definition}

One common algorithmic approach for achieving differential privacy is adding noise that scales with the \emph{sensitivity} of the function being evaluated, which is the maximum change in the function's value that can result from changing a single data point. Differentially private algorithms are robust to \emph{post-processing}, meaning that any data-independent function of a differentially private output retains the same privacy guarantee, and they enjoy \emph{composition}, meaning that the privacy parameters degrade gracefully as additional analyses are performed on the dataset. The simplest version of composition is that the privacy parameters $\varepsilon$ and $\delta$ ``add up'' over multiple analyses, although stronger versions of composition are also used.

Differentially Private Stochastic Gradient Descent (DP-SGD, presented formally in Algorithm \ref{alg:dp-sgd} in Appendix \ref{app.dpsgd}) was introduced by \citep{Abadi_2016} as a method for private non-convex optimization. At each step $t$, DP-SGD subsamples\footnote{The original algorithm of \citep{Abadi_2016} does this via Poisson subsampling, but it can also be done via uniform subsampling while retaining a privacy guarantee \citep{DBLP:journals/corr/abs-1808-00087}.} a small set of data points and uses this batch to compute a gradient update. To achieve a differential privacy guarantee, DP-SGD adds mean-zero Gaussian noise to the average of the per-example gradients. The standard deviation of this noise is scaled with the sensitivity of the gradient estimation. Since this is unbounded, the per-example gradients are first clipped to ensure that the $\ell_2$-norm is at most some input parameter $C$, thus bounding the sensitivity, and then adds noise which scales with $C$.

\citep{Abadi_2016} also introduced the \emph{moments accountant}, which provides tight privacy composition across multiple gradient update steps in DP-SGD. To describe the moments accountant, given an algorithm $\mathcal{M}$ and two neighboring datasets $D, D'$, first we denote the privacy loss of a particular outcome $o$ as $L^{(o)} = \log (\Pr(\mathcal{M}_\mathcal{D} = o) / \Pr(\mathcal{M}_\mathcal{D'} = o))$. The moments accountant calculates a privacy budget by bounding the moments of the privacy loss random variable $L^{(o)}$. That is, if we consider the log of the moment generating function (MGF) of the privacy loss random variable evaluated at $\lambda$, i.e., $\alpha_{\mathcal{M}}(\lambda; \mathcal{D}, \mathcal{D'}) = \log \mathbb{E}_{o \sim \mathcal{M}_{\mathcal{D}}}[e^{\lambda L^{(o)}}]$, the worst case over all neighboring databases $\max_{\mathcal{D}, \mathcal{D'}} \alpha_{\mathcal{M}}(\lambda; \mathcal{D}, \mathcal{D'})$ composes linearly across multiple mechanisms (Theorem 2.1 \citep{Abadi_2016}) and allows for conversion to an associated $(\varepsilon, \delta)$-differential privacy guarantee through the relation $\delta = \min_{\lambda} \exp [\alpha_{\mathcal{M}}(\lambda) - \lambda \varepsilon]$. Follow up work of \citep{bu2019deep} introduced NoisySGD, which followed the same algorithmic structure but analyzed privacy composition under Gaussian differential privacy \citep{DBLP:journals/corr/abs-1905-02383}. For the purpose of this work it is sufficient to simply note the associated benefits of analysis under Gaussian differential privacy: it naturally lends itself to composition under subsampling, allows for analytically tractable expressions of the privacy guarantees of NoisySGD, while providing a slightly tighter overall privacy bound than that achieved by the moments accountant. Further details are provided in Appendix \ref{app.prelims}.

\section{Differentially Private Normalizing Flows}\label{s.algo}

In this section we introduce our algorithm for differentially private density estimation via normalizing flows, DP-NF, presented in Algorithm \ref{alg:DP-NF}. It is based on the DP-SGD algorithm of \citep{Abadi_2016}, which is a differentially private method for performing stochastic gradient descent. We also briefly discuss performance improvements using data-dependent initialization of normalization layers and using a differentially private estimate of the distribution to act as a prior, both of which are explored further Appendix \ref{app.extend}. We emphasize that our primary technical contribution is not in the design of these algorithms, but rather the novel application of these tools to the problem of differentially private density estimation in a way that yields substantial performance over prior work, as demonstrated by our empirical results in Section \ref{s.exp}.

\subsection{DP-NF Algorithm}
Training a normalizing flow model corresponds to minimizing the loss function in Equation \eqref{eq.loss}: $\mathcal{L}(\bm{\theta}) = -\frac{1}{N} \sum_{i=1}^{N} \log p_{\bm{\theta}}(\bm{x}^{(i)})$.  This loss function is non-convex when applied to the optimization of a non-linear normalizing flow model, and hence optimization is typically performed via gradient descent on $\bm{\theta}$. To make this training private in Algorithm \ref{alg:DP-NF}, we update $\bm{\theta}$ using the DP-SGD algorithm of \citep{Abadi_2016} described in Section \ref{s.dp}, with some subtle yet important augmentations to the standard minibatch gradient descent process to allow for an explicit privacy guarantee, in accordance with DP-SGD.

\begin{algorithm}[H]
\caption{DP-NF, Differentially private density estimation via normalizing flows}
\label{alg:DP-NF}
\begin{algorithmic}[1]
\STATE \textbf{Input:} Dataset $\bm{X} = \{\bm{x}^{(1)}, \ldots, \bm{x}^{(n)}\}$, initialized parameters $\bm{\theta}$, learning rate $\eta$, batch size $b$, noise scale $\sigma$, upper-bound on $\ell_2$ norm of per-example gradient $C$, training privacy budget $\varepsilon$, training privacy tolerance $\delta$, privacy accountant $P$.
\STATE $t \gets 1$
\WHILE{$P(t, b / n, \sigma, C, \delta) < \varepsilon$}
    \STATE Take a uniformly random subsample $I_t \subseteq \{1, \ldots, n\}$ with batch size $b$.
    \FOR{$i \in I_t$}
        \STATE $\bm{g}_t^{(i)} \gets \nabla_{\bm{\theta}} -\log p_{\bm{\theta}}(\bm{x}^{(i)})$
        \STATE $\bm{\bar{g}}_t^{(i)} \gets \bm{g}_t^{(i)} / \max \{ 1, || \bm{g}_t^{(i)} ||_2/C \}$
    \ENDFOR
    \STATE $\bm{\theta} \gets \bm{\theta} - \eta \cdot \frac{1}{b}(\sum_i \bm{\bar{g}}_t^{(i)} + \mathcal{N}(\bm{0}, \sigma^2C^2\bm{I}))$
    \STATE $t \gets t + 1$
\ENDWHILE
\STATE \textbf{Output} $\bm{\theta}$
\end{algorithmic}
\end{algorithm}

First, batches are sampled via uniform subsampling (Line 4). That is, each possible batch of size $b$ has equal likelihood of being chosen (as opposed to repeatedly shuffling the dataset and taking equally sized partitions of the dataset, which is often preferred in practice). Second, rather than computing the gradient with respect to the entire batch, the gradient with respect to each individual data point is calculated, clipped to have maximum $\ell_2$ norm $C$, averaged, then added with a randomly sampled Gaussian noise vector (Lines 6-9).

Algorithm \ref{alg:DP-NF} also requires a \emph{privacy accountant} to be specified as input.  This privacy accountant will dynamically track the $\varepsilon$ privacy loss incurred by composition over all gradient update steps as a function of the training parameters, and will halt the algorithm once a pre-specified budget is reached. A privacy accountant $P(t, b / n, \sigma, C, \delta)$ takes in the round $t$ of training, the sampling probability $b/n$ of a single point (here a batch of size $b$ is sampled uniformly from a set of $n$ data points), the noise scale $\sigma$ that is added to preserve privacy, the bound $C$ on the $\ell_2$ norm of each gradient, and the privacy parameter $\delta$. At every time step, the privacy accountant maintains the current $\varepsilon$ privacy budget that has been expended until round $t$ given the input parameters. Common choices for this accountant include the moments accountant (MA) \citep{Abadi_2016} or composition via Gaussian differential privacy (GDP) \citep{DBLP:journals/corr/abs-1905-02383}. In our experiments in Section \ref{s.exp}, we yield preferable results using a GDP privacy accountant.

In summary, DP-NF in Algorithm \ref{alg:DP-NF} is a modified version of DP-SGD, instantiated to train a normalizing flow model with the analyst's choice of privacy accountant.

The privacy guarantees of DP-NF follow as an immediate corollary from those of DP-SGD \citep{Abadi_2016} when instantiated with the moments accountant, and from NoisySGD \citep{bu2019deep} when instantiated with the Gaussian differential privacy accountant.

\begin{theorem}
DP-NF is $(\varepsilon,\delta)$-differentially private.
\end{theorem}

\subsection{DP-NF Extensions}

In practice, one will find that many deep learning models (including the normalizing flow models used in our experiments) are much better optimized using adaptive learning rate optimization schemes. Given this, we found significant benefit in using a direct extension to DP-SGD which applies noisy gradients to the model according to the Adam \citep{kingma2014adam} optimizer. Both methods achieve identical privacy guarantees given that computation of the first and second moments of the noisy gradients are merely deterministic data-independent functions of them. Thus they differ only in the post-processing of the noisy gradients, and the privacy guarantees are unchanged.

Two further extensions of Algorithm \ref{alg:DP-NF} are proposed below, which may provide substantial improvements to empirical performance.

\textbf{Data-Dependent Initialization of Normalization Layers.} Intermediate normalization layers such as activation normalization \citep{kingma2018glow} have been proposed as a means to improve the stability of normalizing flow models. Activation normalization is characterized by a feature-wise offset and scaling of inputs by a learned set of parameters $\bm{b}$ and $\bm{w}$, i.e., $(\bm{x}^{(i)} - \bm{b}) / \bm{w}$. In practice, these parameters are typically set via data-dependent initialization \citep{DBLP:journals/corr/SalimansK16} by setting $\bm{b}$ and $\bm{w}$ as the per-feature means and standard deviations observed throughout a forward pass of a sampled batch of data.  These parameters can also be estimated privately, e.g., by applying the Laplace Mechanism \citep{DMNS06} to the clipped mean and standard deviation, thus allowing for data-dependent initialization of these normalization layers. For more details, see Appendix \ref{s.norm}.

\textbf{Differentially Private Data-Dependent Priors.} Section \ref{s.normflow} suggested the analyst choose a data-independent prior $q$, such as the multivariate spherical Gaussian. However, recent work suggests that modest improvements in empirical results can be achieved through the use of more complex priors, such as a mixture of Gaussians \citep{NIPS2017_6828}, or by fitting a Gaussian mixture model to the data \citep{izmailov2019semisupervised}. A natural privacy-preserving approach would be to first use DP-MoG \citep{pmlr-v54-park17c} with privacy budget $(\varepsilon_1, \delta_1)$ to estimate a prior, and then refine the prior using DP-NF with privacy budget $(\varepsilon_2, \delta_2)$ to yield an encompassing normalizing flow model. This process would be $(\varepsilon_1 + \varepsilon_2, \delta_1 + \delta_2)$-differentially private, and may yield preferable results in settings where the distribution is highly discontinuous, but also locally non-linear. For more details, see Appendix \ref{s.dpem}.

\section{Experimental Results}\label{s.exp}

In this section we present experimental results demonstrating the empirical performance of our approach, evaluating our algorithm on a variety of real and synthetic datasets on varying tasks. In the main body we focus our evaluation on a single dataset (the Life Science dataset \citep{Dua:2019},  described next), although refer to Appendix \ref{app.results} for all additional results on other real and synthetic datasets.  

In all our experiments on the Life Science dataset, we used $\delta=1.52 \times 10^{-5}$. Our baseline method for comparison \citep{pmlr-v54-park17c} used $\delta=1.00 \times 10^{-4}$. However, this corresponds to $\delta=O(1/n)$ on this dataset, which is typically deemed unacceptably large in the privacy community. Instead, our choice of $\delta = 1.52 \times 10^{-5} = 1/n^{1.1}$, which is sublinear in the size of the database.  Smaller values of $\delta$ would not change our qualitative results, nor would they substantially change our quantitative results.

\subsection{Datasets, Implementation, \& Setup}

The Life Science dataset is a standard density estimation benchmark dataset from the UCI machine learning repository \citep{Dua:2019} containing 26,733 real-valued records of dimension 10. This dataset was used in the original evaluation of our baseline model \citep{pmlr-v54-park17c}. Results using additional datasets are presented in Appendix \ref{app.results}. 

Experiments were run on a machine with 2 CPUs, 13 GB RAM, and a single NVIDIA Tesla K80 GPU, and took on the order of half an hour to five hours to run in wall-time, depending on the number of iterations and the dimensionality of the dataset. Models were implemented in the Jax \citep{jax2018github} deep learning framework, and used privacy accounting implementations from TensorFlow Privacy \citep{tensorflowprivacy}.

\textbf{Hyperparameter Search and Model Selection.} Reported privacy budgets in our results correspond only to the training of each model, and does not include privacy loss from hyperparameter search and model selection. We chose not to select hyperparameters in a privacy-preserving manner because this was not the focus of our contribution and because it was not done in our baseline method.\footnote{These can be done privately. For example, \citep{DBLP:journals/corr/abs-0903-4510} provides discrete optimization methods that can be used for private hyperparameter search over discrete model architectures. \citep{Beaulieu-Jones159756} uses ReportNoisyMax \citep{10.1561/0400000042} for private model selection. Some work has also been done to account for high-performance
models without having to spend a significant privacy budget \citep{NIPS2013_5014, DBLP:journals/corr/abs-1811-07971}.} It was generally observed that choices in network structure itself had relatively negligible impacts on results. We found that training parameters such as the gradient clipping bound and batch size had a much more substantial impact on model performance, which is consistent with observations made in \citep{Abadi_2016}.

\textbf{Model Architecture.} The architecture of the model used in our experiments was a variant of a Masked Autoregressive Flow (MAF) \citep{NIPS2017_6828} composed of a repeated sequence of five blocks, each containing a MADE \citep{DBLP:journals/corr/GermainGML15} layer, a reversal layer, and an optional activation normalization layer. Models were optimized via Adam, with default parameters of $\beta_1 = 0.9$ and $\beta_2 = 0.999$. Further details of training parameters and procedures are given in Appendix \ref{app.training}.

\subsection{Empirical Performance of DP-NF}

We implemented our algorithm for differentially private normalizing flows on the Life Science dataset (and other datasets as described in Appendix \ref{app.results}), and evaluated our performance against the baseline of DP-MoG \citep{pmlr-v54-park17c} for a variety of quantitative and qualitative metrics related to density estimation tasks.

\begin{figure}[h!]
\centering
\includegraphics[width=0.6\linewidth]{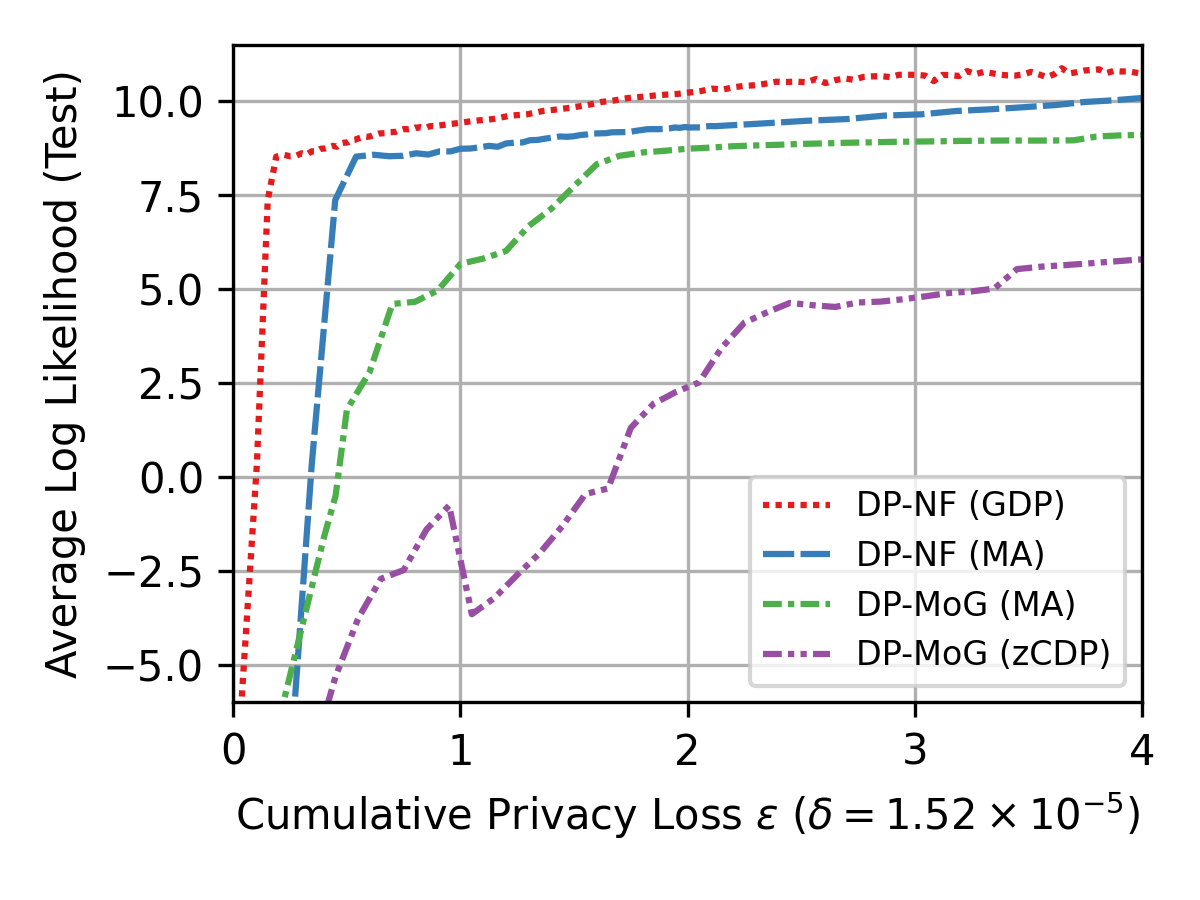}
\caption{Average log likelihood (higher is better) across ten independent cross-validation splits as a function of the cumulative privacy loss $\varepsilon$. DP-MoG was configured to use the Gaussian mechanism with 3 components, as per the original work. DP-NF composed with GDP (as well as MA for fair comparison).}
\label{fig:lifesci-likelihoods}
\end{figure}

\subsubsection{Quantitative Evaluation: Expected Log Likelihood.}
Arguably the most foundational metric for density estimators is the expected log likelihood they assign to held out test points. Figure \ref{fig:lifesci-likelihoods} presents average log likelihood assigned to a held out test set under DP-NF and the baseline method DP-MoG \citep{pmlr-v54-park17c} as a function of $\varepsilon$. We divided the dataset into 10 pairs of training (90\%) and test sets (10\%), and reported the average test log likelihood per data point across the 10 independent trials. Better methods should assign higher log likelihood for points in the held out test set since these points were indeed sampled from the underlying distribution of interest. We found that DP-NF reliably assigned much higher likelihoods to holdout data than that of DP-MoG for identical privacy budgets, across a variety of privacy accountant methods.


The privacy guarantees of DP-NF proved quite practical, providing substantial privacy improvements over DP-MoG for the same model performance. For example, DP-NF matched the peak performance of DP-MoG (achieved around $\varepsilon \approx 4$) for only an expenditure of $\varepsilon \approx 0.5$.
These results are also listed in Table \ref{fig:likelihoods} with error bars showing standard deviation across 10 independent runs.

\begin{table*}[h!]
  \caption{Average test log likelihood for varying privacy budgets $\varepsilon$. Error bars denote standard deviation over ten independent cross-validation splits. Bolded results denote best performing model for a given $\varepsilon$.}
  \centering
  \begin{tabular}{lrrrrr}
    \toprule
        \textbf{Life Science} \\
        $\delta=1.52 \times 10^{-5}$ & $\varepsilon = 0.50$ & $\varepsilon = 1.00$ & $\varepsilon = 2.00$ & $\varepsilon = 4.00$ \\
        \midrule
        DP-NF (GDP)  & \bm{$8.90 \pm 0.18$} & \bm{$9.41 \pm 0.12$} & \bm{$10.20 \pm 0.09$} & \bm{$10.77 \pm 0.24$} \\
        DP-NF (MA)   &     $7.37 \pm 0.17$ &      $8.67 \pm 0.12$ &      $9.32 \pm 0.09$ &      $10.09 \pm 0.18$ \\
        \midrule
        DP-MoG (MA)   &     $2.30 \pm 0.27$ &      $5.09 \pm 0.20$ &       $8.87 \pm 0.06$ &      $9.10 \pm 0.06$ \\
        DP-MoG (zCDP) &    $-8.93 \pm 0.49$ &     $-0.16 \pm 0.37$ &       $2.95 \pm 0.28$ &      $5.48 \pm 0.18$ \\
    \bottomrule
  \end{tabular}
  \label{fig:likelihoods}
\end{table*}

\citep{pmlr-v54-park17c} showed performance of DP-MoG under several different privacy accountant methods, with the moments accountant of \citep{Abadi_2016} providing the best performance. We compared DP-NF using the moments accountant for fair comparison, and using the novel Gaussian differential privacy (GDP) accountant of \citep{bu2019deep}.  Figure \ref{fig:lifesci-likelihoods} and Table \ref{fig:likelihoods} show that DP-NF outperforms DP-MoG for all privacy accountant methods considered for either model, emphasizing that while the GDP accountant does provide some benefit, the vast majority of the performance improvements come from the DP-NF method itself. The benefits of using the GDP accountant are further explored in the appendix.

\subsubsection{Quantitative Evaluation: Downstream Machine Learning Tasks.}

Next we further evaluate the quality of our model by measuring the performance of downstream
machine learning models trained on its generated synthetic data. A proper method for evaluating the strength of density estimation approaches is through the quality of their synthetic data, as measured by the ability to train a machine learning model that performs well on future, real data.


To perform this evaluation, we trained DP-NF and DP-MoG (along with their non-private variants for reference) to learn the distribution of the training data, and then used these models to generate a synthetic dataset. We then trained a simple regressor---$k$-nearest neighbors with default library settings ($k=3$)---on each synthetically generated dataset and evaluated their in predicting a target value on real, held out Life Science data. The Life Science dataset does not have an immediately associated prediction task, as it is primarily used as solely a density estimation benchmark. To artificially construct a prediction task, we simply chose to isolate the last column to act as a label and treated the remaining nine columns as features.

\begin{figure}[h]
\centering
    \includegraphics[width=0.6\textwidth]{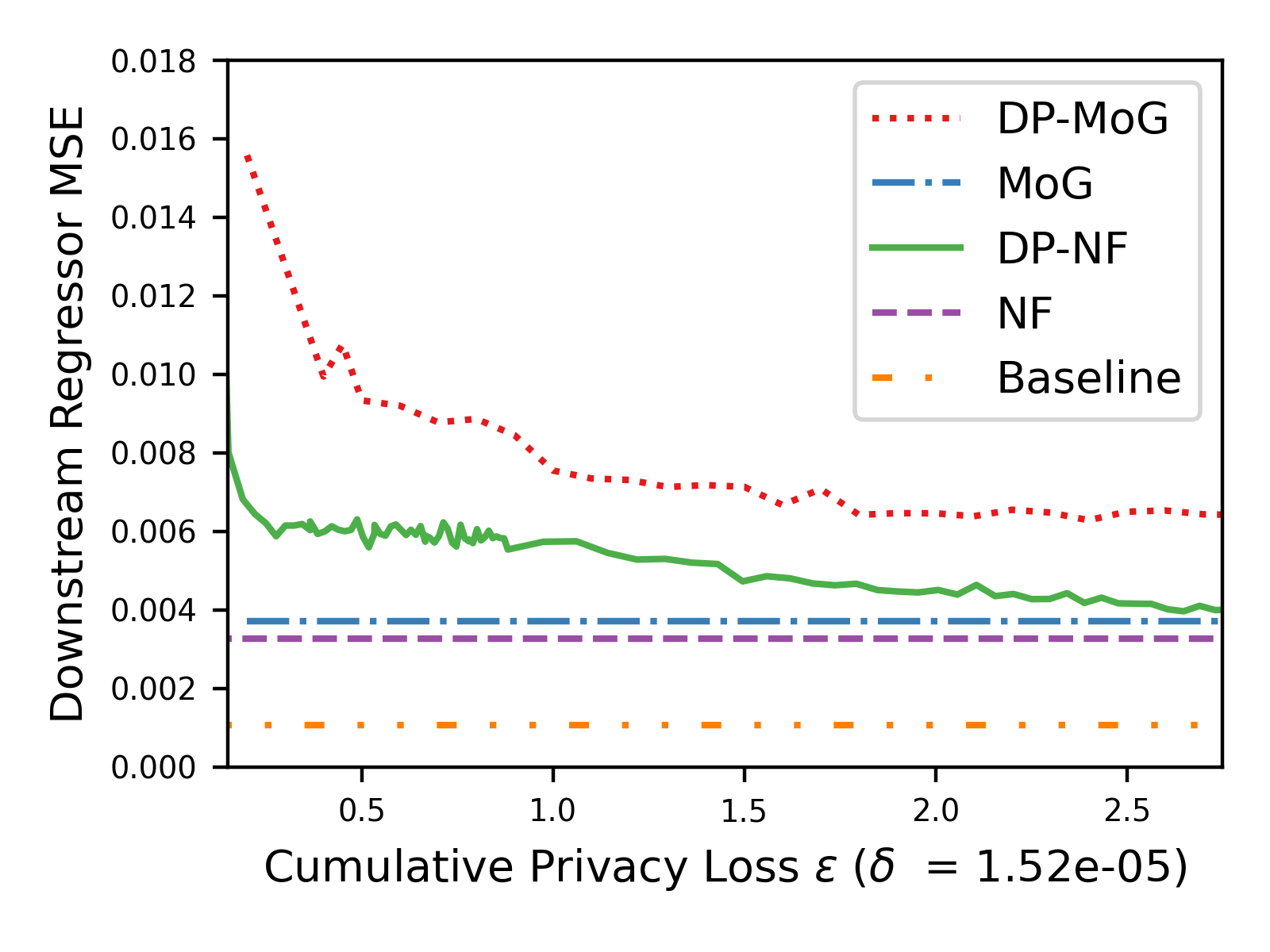}
    \caption{Mean squared error of regressor ($k$-nearest neighbors with $k=3$) on real, held out test data when trained on synthetically generated data by various approaches. Baseline refers to training the regressor on the real data. }
    \label{fig:downstream-task}
\end{figure}

Figure \ref{fig:downstream-task} shows the mean squared error attained by each regressor as a function of the privacy expenditure of the data generation approach. Horizontal lines denote that the approach is $\varepsilon=\infty$, i.e., non-private. ``Baseline'' refers to training the regressor directly on the training dataset provided to the density estimator. Upon inspection, we find that DP-NF generates data of convincingly higher quality than that of our comparison method DP-MoG for all values of $\varepsilon$. In addition, for higher values of $\varepsilon$, DP-NF converges to the quality achievable by a non-private MoG, whereas DP-MoG hits an apparent plateau well before this point.

\subsubsection{Qualitative Evaluations.}

\begin{figure}[H]
\centering
\includegraphics[width=0.55\linewidth]{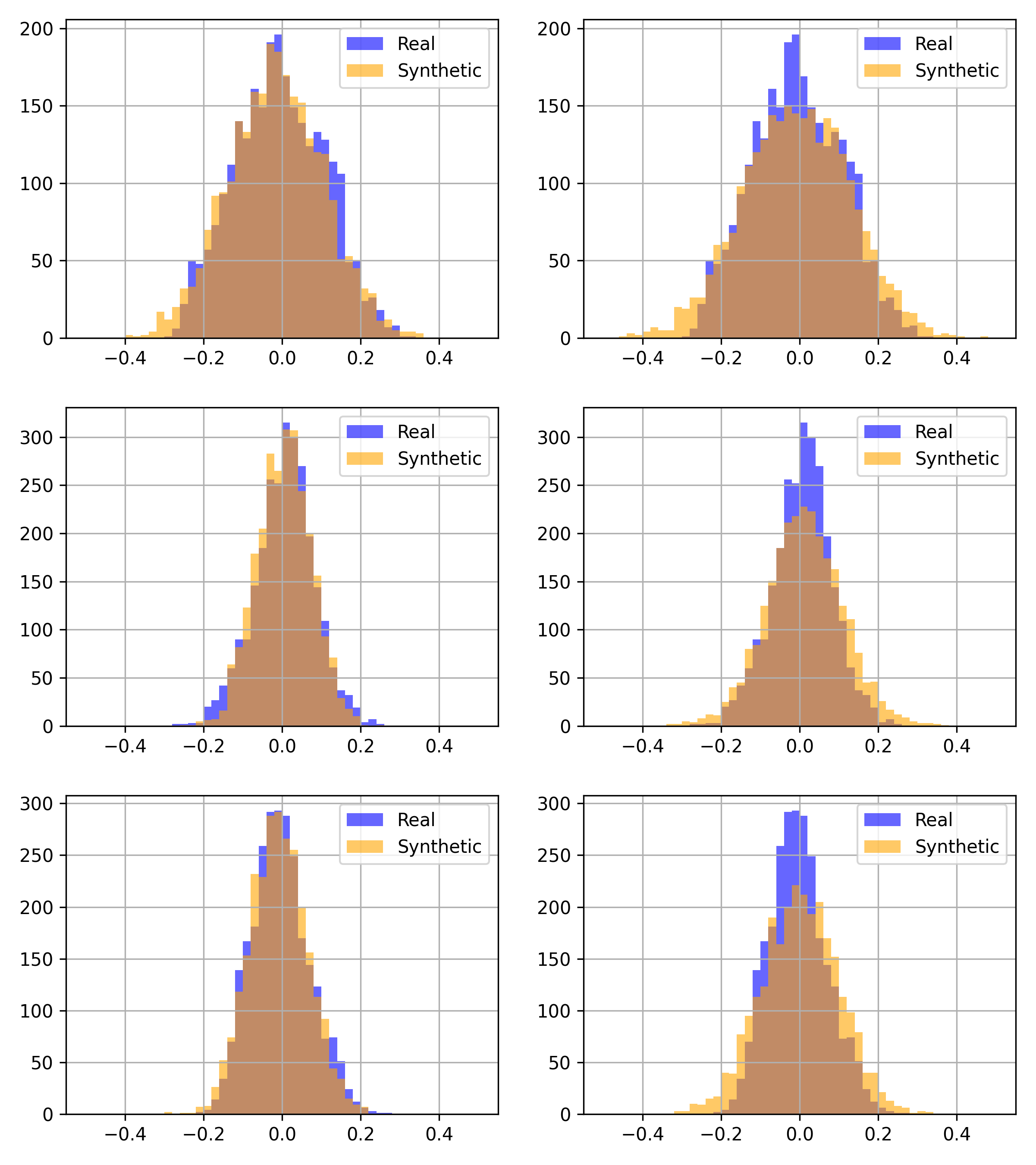}
\caption{Dimension-wise histograms of synthetically generated Life Science data, superimposed over real data, for $\varepsilon = 0.6$ and $\delta = 1.52 \times 10^{-5}$. \textbf{Left Column:} DP-NF. \textbf{Right Column:} DP-MoG. Note DP-NF's ability to capture regions of concentrated density, whereas DP-MoG struggles in this respect.
}
\label{fig:lifesci-samples-dimwise-nf}
\end{figure}

Figure \ref{fig:lifesci-samples-dimwise-nf} shows that DP-NF provides a qualitative increase in sample quality under visualization. It presents dimension-wise histograms of synthetically generated features for three features of the Life Science dataset, using DP-NF (left column) and DP-MoG (right column) for comparison. (See Figure \ref{fig:lifesci-samples-dimwise} in Appendix \ref{app.results} for dimension-wise histograms of all 10 features.)  Both methods used $\varepsilon=0.6$ and $\delta = 1.52 \times 10^{-5}$. In every plot, the synthetic data in orange is superimposed over the real data in blue.  We qualitatively see that for nearly all ten features, the distribution of data generated  by DP-NF closely matches that of the real data, while DP-MoG was relatively unable to replicate regions of concentrated density for certain dimensions. This could be due to the fact that that for a fixed number of components, the DP-MoG model is constrained to cover the support of the distribution and must ignore nuanced details. Normalizing flow models, on the other hand, have heightened expressiveness over traditional statistical methods like Gaussian mixture models, and we see that they are able to capture these nuances more readily.

\begin{figure}[H]
    \centering
    \includegraphics[width=\linewidth]{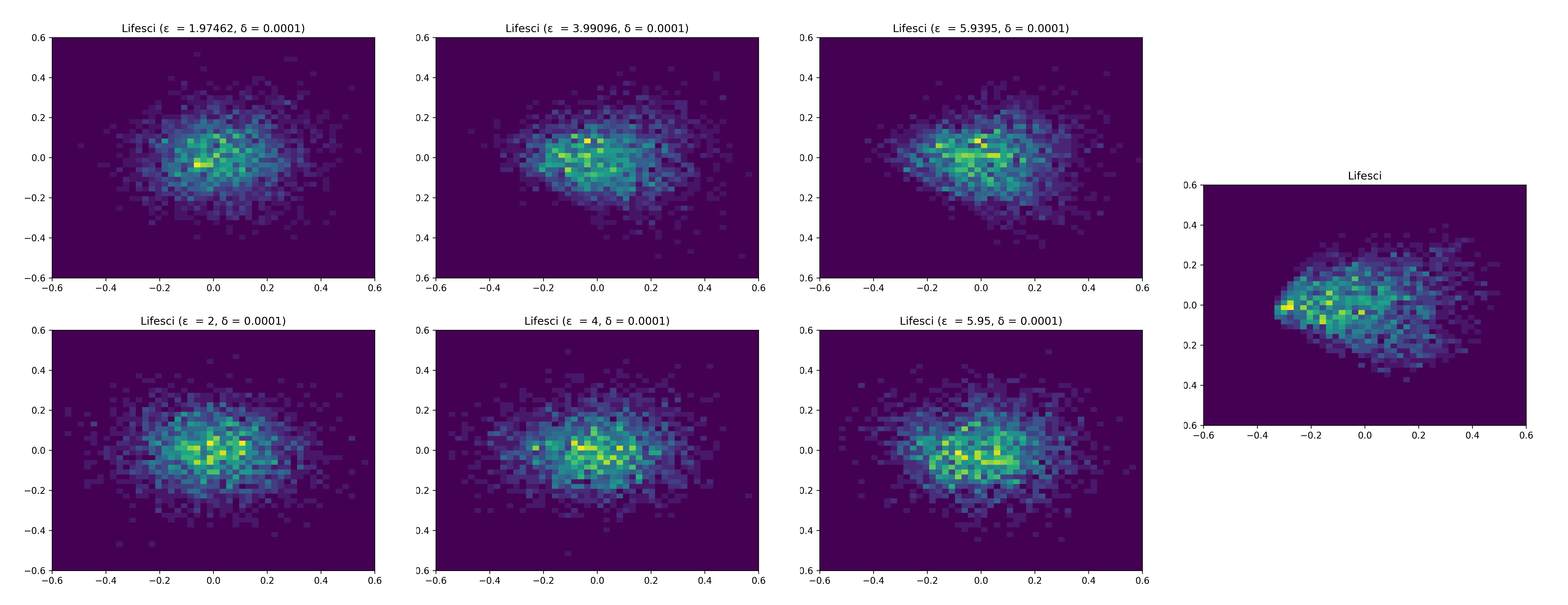}
    \caption{Synthetically generated Life Science data for $\varepsilon$ = 2, 4, and 6, projected to two dimensions via PCA. \textbf{Top row:} DP-NF. \textbf{Bottom row:} DP-MoG. \textbf{Right:} Real data. Note the compression to the left of the distribution of real data that is captured by DP-NF as $\varepsilon$ increases, but not present in the synthetic data generated by DP-MoG.}
    \label{fig:lifesci-samples}
\end{figure}

As another qualitative evaluation of sample visualization, Figure \ref{fig:lifesci-samples} shows the density of synthetic data generated by each model when projected to two dimensional space via PCA, for varying $\varepsilon$ values. The top row shows DP-NF, the bottom row shows DP-MoG, and the right figure shows the real data. In all plots, lighter pixels correspond to regions of higher density, and dark pixels indicate lower density. We see that DP-NF is better able to capture some of the observable qualities exhibited in the real data, for example the gradual compression of density to the left of the distribution.

\section{Application: Differentially Private Anomaly Detection}\label{s.anomaly}

Our DP-NF algorithm can be used as a tool for differentially private anomaly detection.  Given a density estimator, a straightforward approach to anomaly detection is through a simple likelihood thresholding mechanism. For a given point, to determine whether it is in-distribution or out-of-distribution, we can simply return a binary value which denotes in-distribution if the log likelihood assigned to the point by the model is above some empirically derived threshold $T$, and out-of-distribution otherwise. For the purposes of our experiments, we assume that such a threshold is easily estimated, i.e., by selecting the value for $T$ which optimizes anomaly detection performance on a public test set.  In the private setting, we can approach this task in a privacy-preserving manner by training either DP-NF or DP-MoG on the dataset. By the post-processing property of differential privacy, we can make arbitrarily many anomaly detection queries to the privately trained model while incurring no privacy loss beyond what is incurred during training. 

\begin{figure}[h!]
\centering
\includegraphics[width=0.6\linewidth]{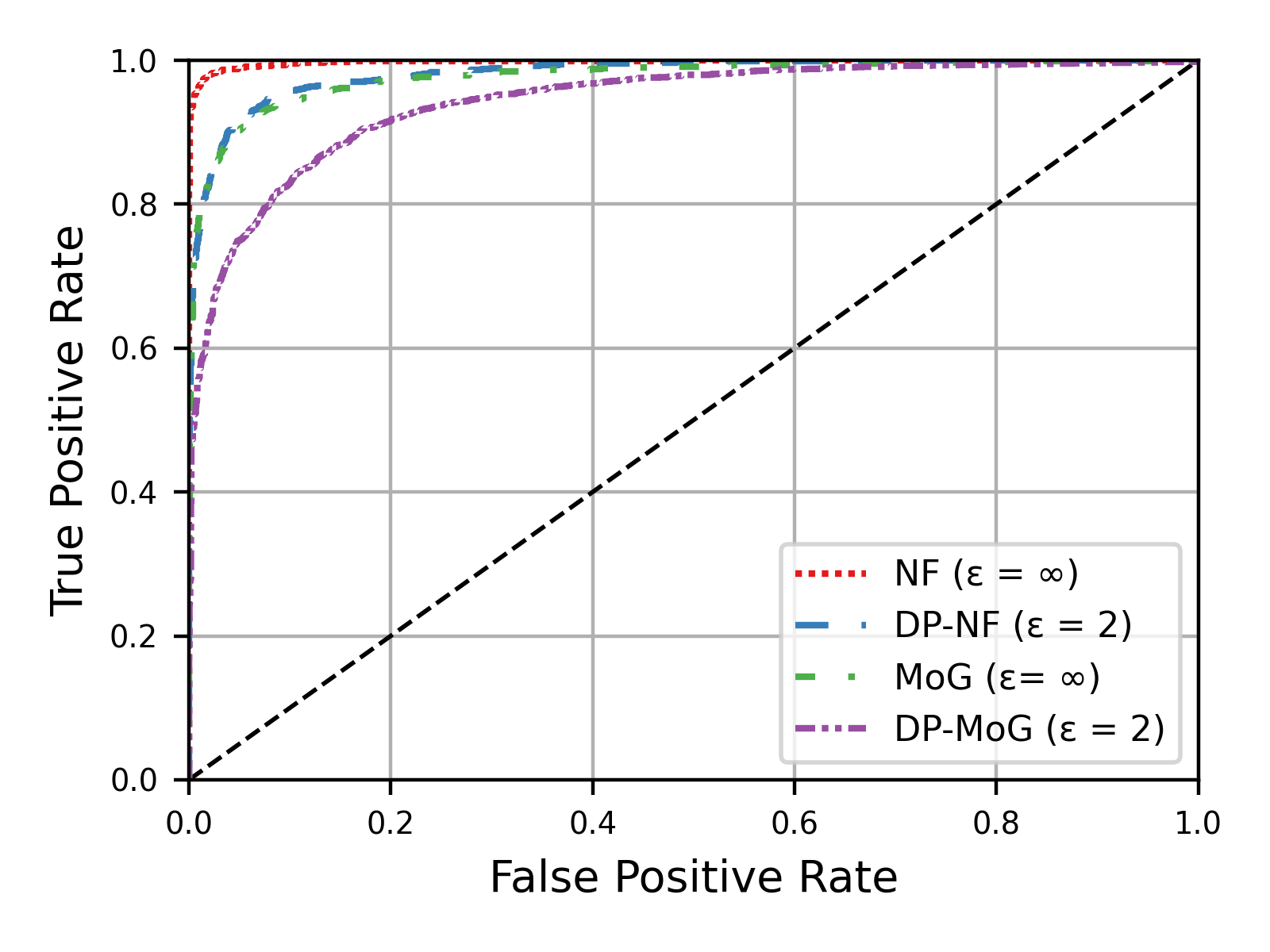}
\caption{ROC curves displaying true positive rate and false positive rate for private and non-private likelihood threshold models. Privacy expenditure was calculated using the moments accountant with $\delta= 1.52 \times 10^{-5}$.}
\label{fig:lifesci-anomaly-detection}
\end{figure}

In Figure \ref{fig:lifesci-anomaly-detection}, we illustrate the efficacy of our approach in performing anomaly detection under this likelihood thresholding mechanism. We randomly generated points that were uniformly distributed around the tails of the test dataset, i.e., between the 5th and 30th percentiles and the 70th and 95th percentiles dimension-wise. The total number of synthetically generated anomalies was equal to the total number of test points. Figure \ref{fig:lifesci-anomaly-detection} shows ROC curves of both private and non-private methods for this binary prediction problem. That is, it shows the tradeoff in the true positive rate and the false positive rate in predicting in-distribution or out-of-distribution correctly for varying selections of the likelihood threshold. We observe that DP-NF outperformed the other private method DP-MoG for the same privacy guarantee.  Our approach performed comparably to non-private MoG, and is of course upper-bounded by a non-private normalizing flow.

\begin{algorithm}[H]
\caption{DP-AD, Differentially private anomaly detection via an ensemble of density estimators}
\label{alg:DP-NF-Ensemble}
\begin{algorithmic}[1]
\STATE \textbf{Input:} Dataset $\bm{X} = \{\bm{x}^{(1)}, \ldots, \bm{x}^{(n)}\}$, example $x$, number of partitions $k$, likelihood threshold $T$, privacy budget $\varepsilon$.
\STATE $X_1, \ldots, X_k \gets partition(X, k)$
\STATE $\theta_1, \ldots, \theta_k \gets train(X_1, \ldots, X_k)$
\STATE $c \gets 0$
\FOR{$i \in [k]$}
    \IF{$p_{\theta_i}(x) > T$}
        \STATE $c \gets c + 1$
    \ENDIF
\ENDFOR
\STATE Sample \textbf{response} as ``in-distribution'' with probability $\frac{\exp(\varepsilon c/2)}{\exp(\varepsilon c/2) + \exp(\varepsilon (k-c)/2)}$ and ``out-of-distribution'' with probability $\frac{\exp(\varepsilon (k-c)/2)}{\exp(\varepsilon c/2) + \exp(\varepsilon (k-c)/2)}$
\STATE \textbf{Output response}
\end{algorithmic}
\end{algorithm}

\begin{wrapfigure}{r}{0.5\linewidth}
\centering
\includegraphics[width=0.9\linewidth]{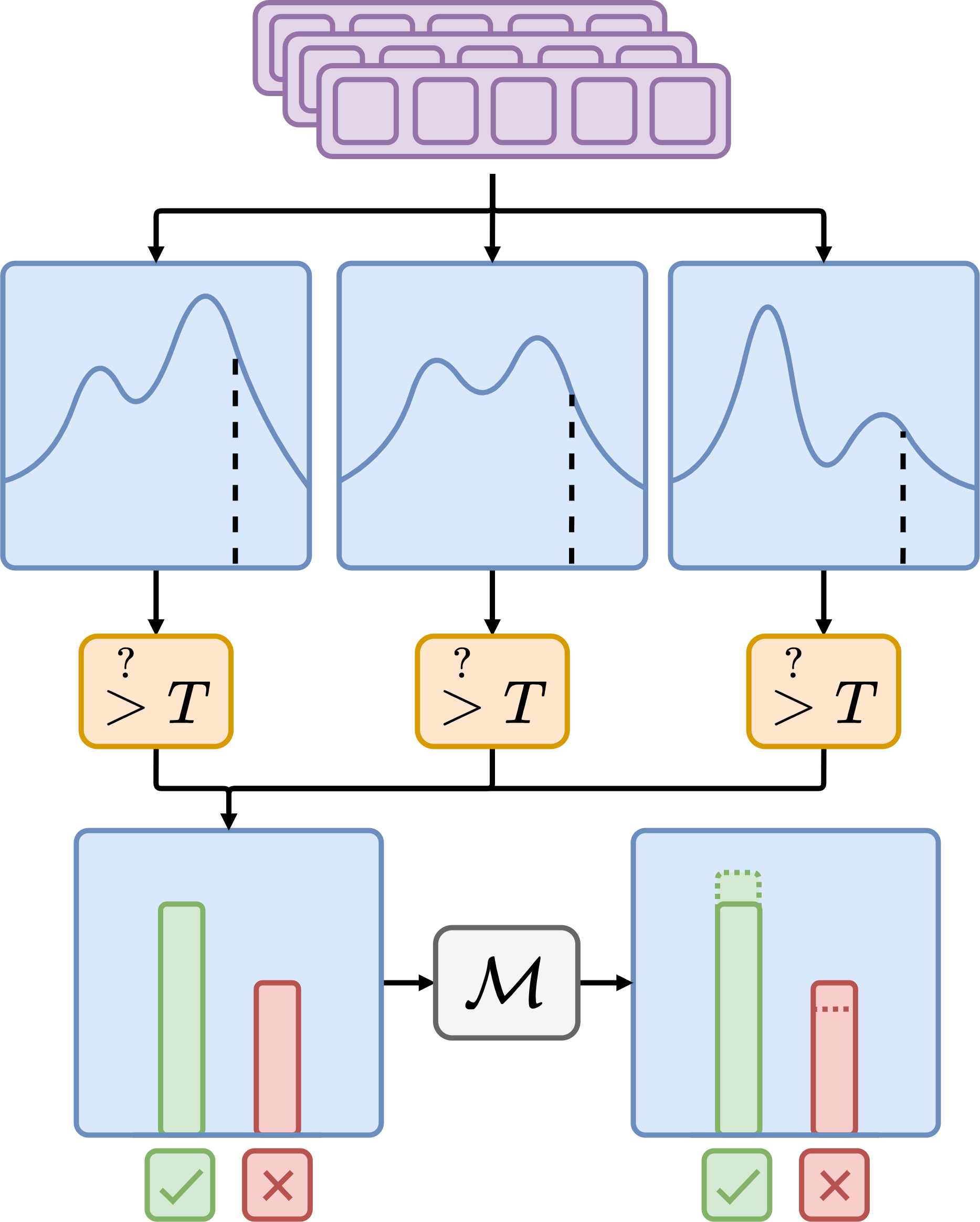}
\caption{Ensembled anomaly detection framework. In this setting, $k$ non-private density estimators are trained on separate partitions of the training dataset. At inference time new examples (purple) are fed to each density estimator, voting on whether the given example is ``in-distribution`` or ``out-of-distribution`` depending on whether it exceeds a known likelihood threshold. Then, we perform a noisy aggregation on these votes to preserve privacy. }
\label{fig:framework-anomaly-detection}
\end{wrapfigure}

There are other possible methods for making anomaly detection algorithms differentially private, beyond the approach described above of using DP-NF directly to privately train the model. An alternative approach is to, rather than learning the parameters of the model in a differentially private manner, partition the dataset into $k$ parts and train a non-private density estimator on each part. Then, given a new point of interest, each model casts a vote regarding its belief on the point being in-distribution or out-of-distribution by testing whether their density assigned to the point exceeds $T$. We then aggregate these votes privately, to ensure that the final prediction is differentially private with respect to the training set.  In our algorithm for differentially private anomaly detection (Algorithm \ref{alg:DP-NF-Ensemble}), we use the Exponential Mechanism \citep{MT07} for this private aggregation. We note that our overall approach is an instantiation of the \emph{sample-and-aggregate} framework of \citep{10.1145/1250790.1250803}. This approach is visualized in Figure \ref{fig:framework-anomaly-detection} and presented formally in Algorithm \ref{alg:DP-NF-Ensemble}.

The privacy guarantees of DP-AD follow as an immediate corollary from those of sample-and-aggregate \citep{10.1145/1250790.1250803} and the Exponential Mechanism \citep{MT07}.

\begin{theorem}
DP-AD is $(\varepsilon,0)$-differentially private.
\end{theorem}

Although Algorithm \ref{alg:DP-NF-Ensemble} is instantiated with the Exponential Mechanism \citep{MT07}, one can opt for any differentially private aggregation method. In contexts requiring answers to a large but bounded number of queries where the number of anomalies is expected to be small in comparison to the total number of queries (an appropriate assumption for many applications), an immediate extension of this approach using the Sparse Vector technique \citep{DNRRV09,DNPR10} would be natural.

\begin{figure}[h!]
\centering
    \includegraphics[width=0.3\columnwidth]{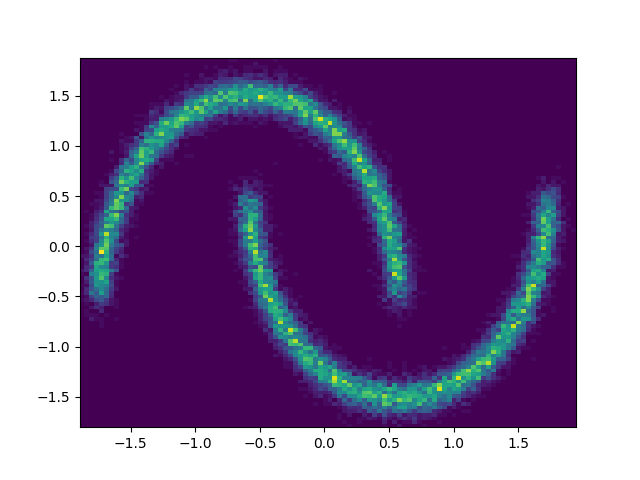}
    \caption{Half-Moons Dataset}\label{fig:moons}
\end{figure}

\begin{figure}[h!]
\centering
  \begin{subfigure}{0.45\columnwidth}
  \centering
  \includegraphics[width=\textwidth]{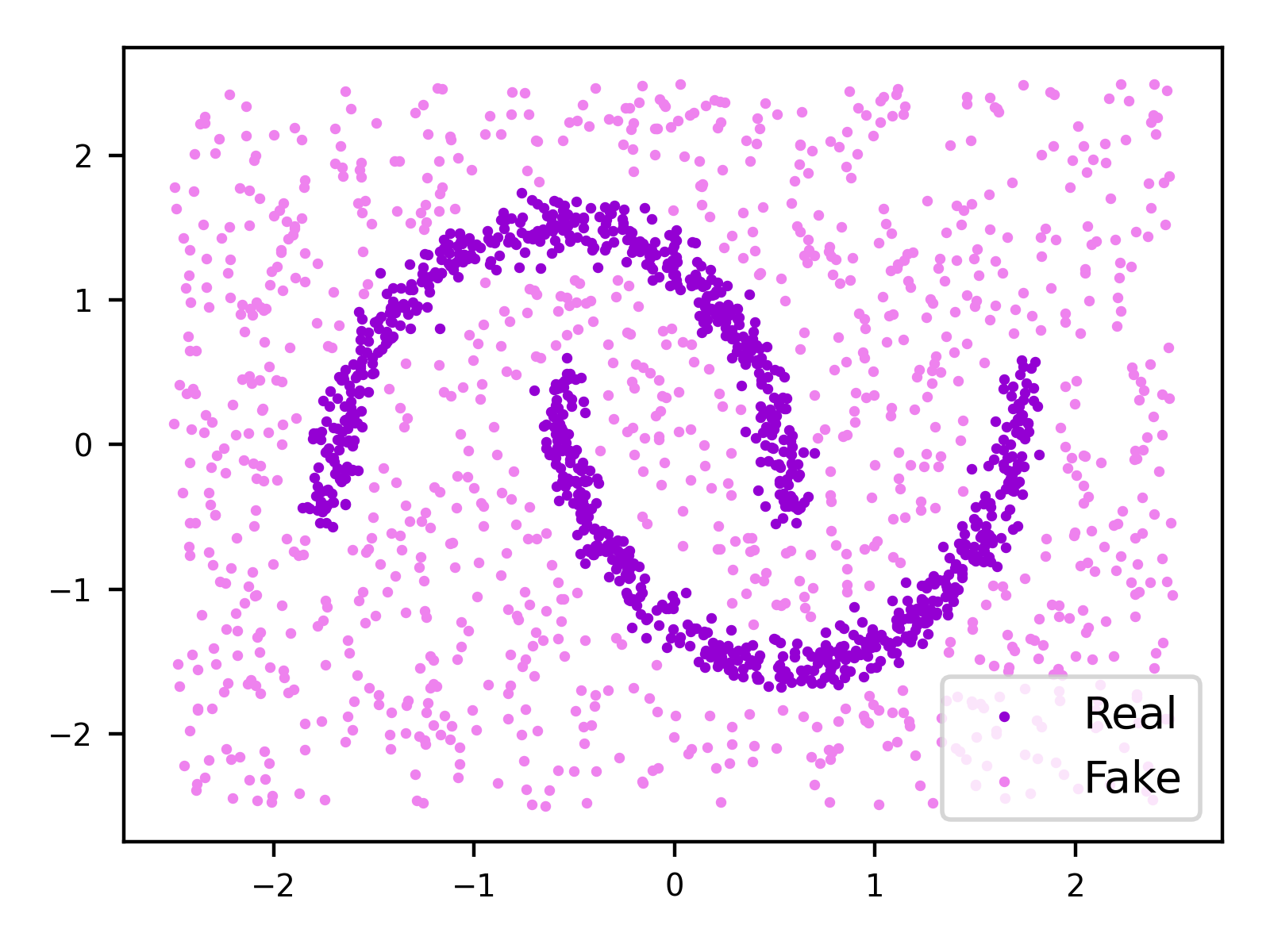}
  \end{subfigure}
  \hfill
  \begin{subfigure}{0.45\columnwidth}
  \centering
  \includegraphics[width=\textwidth]{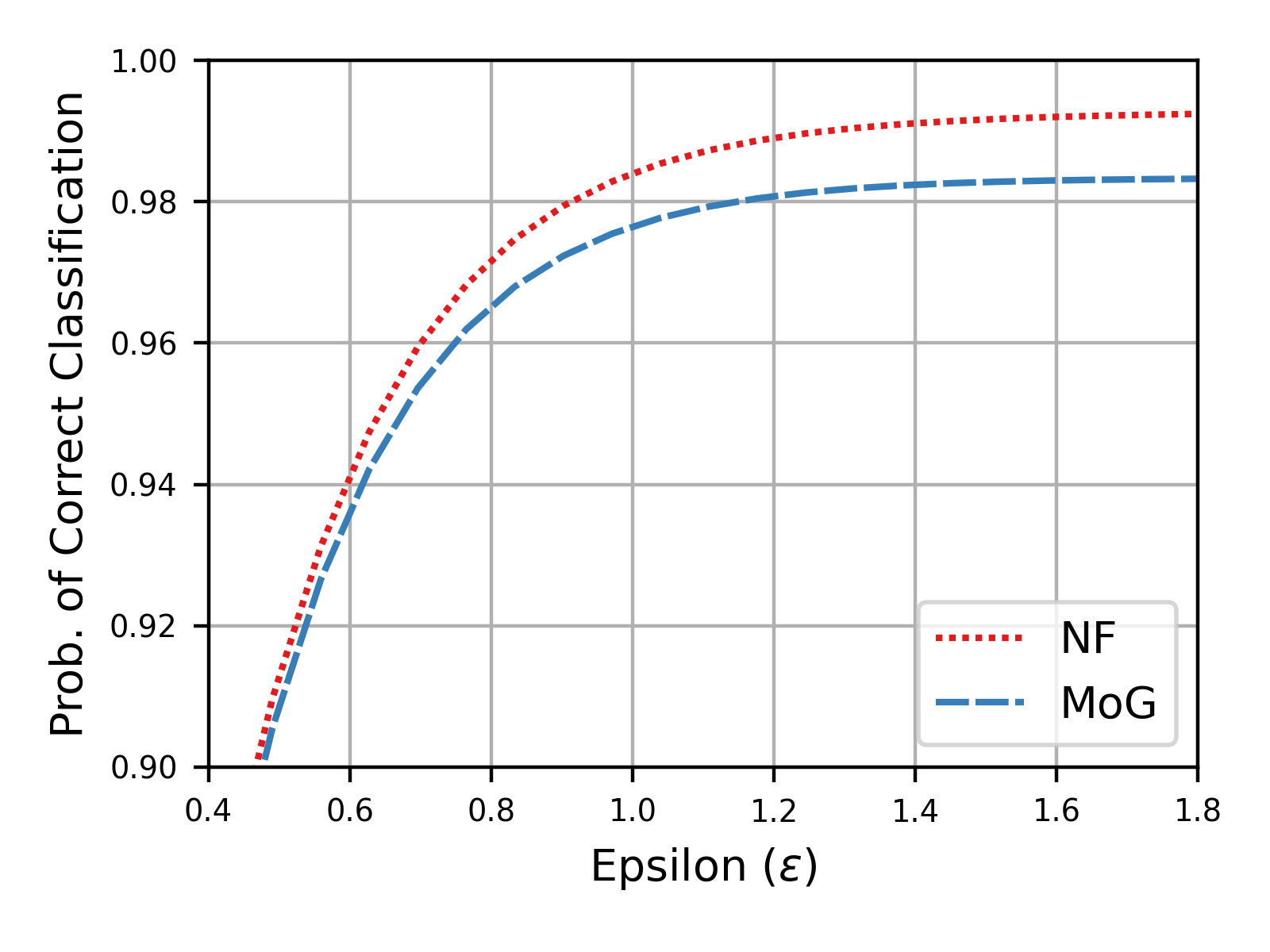}
  \end{subfigure}
\caption{Anomaly detection through private aggregation of votes (either ``in-distribution'' or ``out-of-distribution'') on synthetically generated data. Training set was partitioned into 10 pieces, then each piece was fit to a non-private model, either MoG or NF. Differentially private aggregation was performed via the exponential mechanism.}
\label{fig:anomaly-detection-moons}
\end{figure}

To evaluate this approach, we applied Algorithm \ref{alg:DP-NF-Ensemble} to the Half-Moons dataset, which is a synthetic dataset with 30,000 data points, each with two real-valued features.  This dataset is visualized in Figure \ref{fig:moons}, and more details about this synthetic dataset are given in Appendix \ref{app.results}.

Figure \ref{fig:anomaly-detection-moons} illustrates the data used and the results of this evaluation.  The training data (dark purple) was partitioned into 10 pieces and used to fit a set of independent non-private models (all either MoG or NF). Then, anomalies (light purple) were added to the dataset, and both real and fake data were fed into the learned models. These data are illustrated in the top image of Figure \ref{fig:anomaly-detection-moons}. The Exponential Mechanism was used to perform differentially private aggregation to yield the final ensemble prediction when applied to held out test points. The bottom image of Figure \ref{fig:anomaly-detection-moons} presents the probability of a correct classification between ``in-sample'' and ``out-of-sample'' data as a function of $\varepsilon$ under this scheme. Both approaches were given the threshold that optimized their classification performance. Notice that using normalizing flows in this setting naturally yields a performance improvement over the existing method, which is due in part to the ability of normalizing flows to capture densities which do not adhere to a Gaussian distribution.

The benefit of this ensembled approach is that each individual model can be trained non-privately, which may substantially improve the quality of the learned models. For certain problems, this might be necessary when differentially private optimization is known to degrade performance substantially, for example in high dimensional problems with image data. The associated downside is that the analyst's privacy budget would now degrades as a function of the number of queries, whereas before the privacy budget was independent of the number of queries. The analyst would have to determine the application-specific tradeoff between costs and benefits of this approach, based on practical constraints imposed by the problem context.

\section{Conclusion}

Privacy is a subject of increasing importance and growing concern. By our work adhering to the framework of differential privacy, one is able to make definitive statements regarding the privacy of participants involved in our analysis.  Our results could also be used to enable differentially private synthetic data generation, which would allow data curators to provide privatized synthetic versions of their sensitive or protected datasets, thereby enabling broader access to these data.

In this work, we have demonstrated the efficacy of differentially private normalizing flow models as a novel approach to the task of privacy-preserving density estimation. We have shown the ability of these models to assign high likelihoods to holdout data and generate qualitatively realistic synthetic data, improving on existing state-of-the-art methods. Going forward, there exist several interesting directions for further development. For example, it remains to be seen how normalization layers such as activation normalization, whose parameters are likely disproportionally sensitive to perturbation during differentially private optimization, could be better adapted to such. Further, one might hypothesize that sampling via partitions of a shuffled dataset may yield improved results given more regular sampling of each data point, and associating such a sampling method with rigorous privacy guarantees if possible could yield empirical improvements. Finally, in this study we only considered a particular subset of normalizing flows in existence. Although, many alternative neural density estimators capable of expressing highly discontinuous distributions are in continuous development, including FFJORD \citep{DBLP:journals/corr/abs-1810-01367}, Neural Spline Flows \citep{durkan2019neural}, Neural Autoregressive Flows \citep{DBLP:journals/corr/abs-1804-00779}, and Transformation Autoregressive Networks \citep{oliva2018transformation}. 

%
%
%

\bibliography{references.bib}
\bibliographystyle{plainnat}

\newpage

\appendix


\section{Additional Privacy Preliminaries}\label{app.prelims}

\subsection{Differential Privacy}


We achieve differential privacy in Algorithm \ref{alg:DP-NF} through the Gaussian Mechanism, which adds mean-zero Gaussian noise to the value of a function $f$ evaluated on the data.  The scale of the noise depends on the \emph{sensitivity} of the function.
The $\ell_2$-sensitivity of a function $f$ is denoted $\Delta_2 f$, and is the maximum change in the $\ell_2$ norm of $f$ if one entry in the database were to be changed. Formally, for $f : \mathbb{N}^{|\mathcal{X}|} \rightarrow \mathbb{R}^k$, $\Delta_2 f = \max_{D,D' \text{ neighbors}} \| f(D) - f(D') \|_2$. In our case, this function is the computation of the gradient given a sampled batch of data.





\begin{theorem}[Gaussian Mechanism \cite{10.1561/0400000042}]\label{def.gaussian}
Let $f : \mathbb{N}^{|\mathcal{X}|} \rightarrow \mathbb{R}^k$ and let $\mathcal{M}(x, f, \varepsilon) = f(x) + (Z_1, \ldots, Z_k)$ where $Z_i$ are i.i.d. random variables drawn from $\mathcal{N}(0, \sigma^2)$ and $\sigma > \sqrt{2 \ln(1.25/\delta)} \triangle_2 f / \varepsilon$. Then $\mathcal{M}$ is $(\varepsilon, \delta)$-differentially private.
\end{theorem}

Differential privacy \emph{composes}, meaning that the privacy guarantees degrade smoothly as more analyses are performed on the same dataset. The simplest version of privacy composition is that the $\varepsilon$s and $\delta$s ``add up'' across  analyses.  Tighter composition bounds are possible, including the approaches outlined in the following two subsections.

\subsection{Moments Accountant}





The moments accountant \cite{Abadi_2016} was proposed initially as a means for tight composition of the privacy gaurantees of DP-SGD. To characterize this analysis, we note the privacy loss associated with a given outcome $o$, given as $L^{(o)} = \log \frac{\Pr(\mathcal{M}_\mathcal{D} = o)}{\Pr(\mathcal{M}_\mathcal{D'} = o)}$. Further, for two given datasets we define the MGF of this random variable evaluated at some value $\lambda$ as $\alpha_{\mathcal{M}}(\lambda; \mathcal{D}, \mathcal{D'}) = \log \mathbb{E}_{o \sim \mathcal{M}_{\mathcal{D}}}[e^{\lambda L^{(o)}}]$. Finally, the "worst case" upper bound across all possible pairs of datasets is given as $\alpha_{\mathcal{M}}(\lambda) \triangleq \max_{\mathcal{D}, \mathcal{D'}} \alpha_{\mathcal{M}}(\lambda; \mathcal{D}, \mathcal{D'})$.

This notation can be used to give composition guarantees for the privacy parameters across multiple algorithms run on the same dataset.


\begin{theorem}[\cite{Abadi_2016}]
Suppose that an algorithm $\mathcal{M}$ consists of a sequence of adaptive algorithms $\mathcal{M}_1, \ldots, \mathcal{M}_k$ where $\mathcal{M}_i: \prod_{j=1}^{i-1} \mathcal{R}_j \times \mathcal{D} \rightarrow \mathcal{R}_i$. Then for any $\lambda$, $\alpha_{\mathcal{M}}(\lambda) \leq \sum_{i=1}^{k} \alpha_{\mathcal{M}_i}(\lambda)$.
For any $\varepsilon$ > 0,  $\mathcal{M}$ is $(\varepsilon, \delta)$-differentially private for $\delta = \min_{\lambda} \exp(\alpha_{\mathcal{M}}(\lambda) - \lambda \varepsilon)$.
\end{theorem}

This analysis was then later characterized under the framework of R\'enyi differential privacy \cite{DBLP:journals/corr/Mironov17}, a relaxation of $(\varepsilon, \delta)$-differential privacy which is defined in a manner closely resembling the moments accountant privacy analysis. 



\begin{definition}[($\alpha, \varepsilon$)-RDP \cite{DBLP:journals/corr/Mironov17}]
A randomized algorithm $f: \mathcal{D} \rightarrow \mathcal{R}$ is said to have $\varepsilon$-R\'enyi differential privacy of order $\alpha$, or $(\alpha, \varepsilon)$-RDP for short, if for any adjacent $D, D' \in \mathcal{D}$ it holds that $D_{\alpha}(f(D) || f(D')) \leq \varepsilon$, where $D_\alpha(P || Q) \triangleq \frac{1}{\alpha - 1} \log \mathbb{E}_{x \sim Q} (\frac{P(x)}{Q(x)})^\alpha$.
\end{definition}

Finally, the privacy analysis performed in \cite{Abadi_2016} regarding DP-SGD assumes the sampling is performed via Poisson subsampling, i.e., each individual example has independent sampling probability $q$ of being included in the batch at each iteration. In some contexts, fixed-sized batches can enable a variety of performance improvements by allowing for compilation. Privacy analysis under uniform subsampling, where each batch is sampled uniformly across all possible batches of size $b$, was considered in \cite{DBLP:journals/corr/abs-1808-00087} :

\begin{theorem}[RDP-DP Conversion \cite{DBLP:journals/corr/abs-1808-00087}]
 For all integers $\alpha \geq 2$, if $\mathcal{M}$ is $(\alpha, \varepsilon(\alpha))$-RDP, then the randomized algorithm applied to a subsampled batch of data without replacement is $(\alpha, \varepsilon'(\alpha))$-RDP where $\varepsilon'(\alpha) \leq \frac{1}{\alpha - 1}\log (1 + \gamma^2 {\alpha \choose 2} \min \{ 4(e^{\varepsilon(2)} - 1), e^{\varepsilon(2)} \min\{ 2, (e^{\varepsilon(\infty)} - 1)^2 \} \} + \sum_{j=3}^{\alpha} \gamma^j {\alpha \choose j} e^{(j - 1)\varepsilon(j)} \min \{ 2, (e^{\varepsilon(\infty)} - 1)^j \})$.
\end{theorem}

\subsection{Gaussian Differential Privacy}\label{app.gdp}

Gaussian differential privacy (GDP) is a recently proposed relaxation of $(\varepsilon, \delta)$-differential privacy established in \cite{DBLP:journals/corr/abs-1905-02383}, and further expanded upon in the context of deep learning in \cite{bu2019deep}. This definition exhibits several appealing properties, including simplified analysis under composition and subsampling, derivation of analytically tractable expressions for the privacy guarantees of NoisySGD, while providing a slightly tighter privacy bound than that which is achieved through analysis via the moments accountant. The framework of \emph{$\mu$-Gaussian differential privacy} acts as the basis for our analysis.

We note that the Gaussian mechanism (Definition \ref{def.gaussian}) $\mathcal{M}(D) = f(D) + \mathcal{N}(0, \left(c\Delta_2 f/\varepsilon\right)^2)$ is $(\varepsilon/c)$-GDP \cite{DBLP:journals/corr/abs-1905-02383}.\footnote{Further detail concerning the privacy guarantees achieved when batches are subsampled is given in Section 2 of \cite{bu2019deep}.}

The overall privacy guarantee corresponding to the $k$-fold adaptive composition of $k$ mechanisms each satisfying $\mu_i$-GDP is $\sqrt{\mu^2_1 + \mu^2_2 + \ldots \mu^2_k}$-GDP. Finally, $\mu$-GDP allows for a conversion to a corresponding $(\varepsilon, \delta)$-differential privacy guarantee using the fact that an algorithm is $\mu$-GDP if and only if it is $(\varepsilon, \delta(\varepsilon))$-differentially private for all $\varepsilon \geq 0$, where $\delta(\varepsilon) = \Phi(-\frac{\varepsilon}{\mu} + \frac{\mu}{2}) - e^{\varepsilon} \Phi(-\frac{\varepsilon}{\mu} - \frac{\mu}{2})$ and $\Phi(\cdot)$ is the cumulative density function of the Normal distribution.

\begin{figure}[h!]
\centering
    \includegraphics[width=0.5\columnwidth]{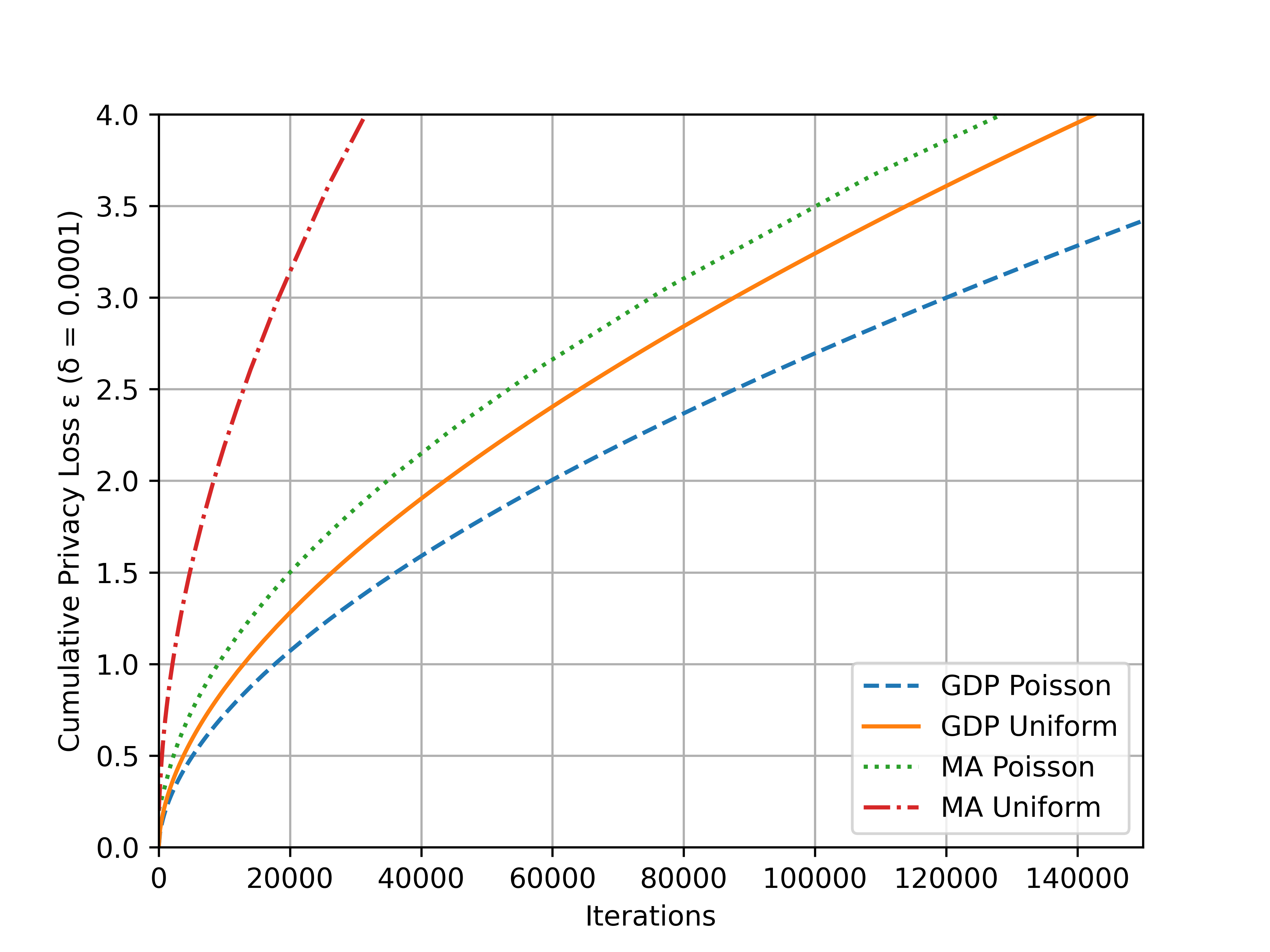}
    \caption{Cumulative privacy loss $\varepsilon$ given Life Science training parameters ($q = b / n = 100 / 21384 = 4.676 \times 10^{-3}, \sigma = 2.1, \delta = 1.00 \times 10^{-4}$) as a function of training iterations.}
    \label{fig:lifesci-privacy-guarantees}
\end{figure}

Figure \ref{fig:lifesci-privacy-guarantees} shows that GDP privacy accounting gives substantial improvements over the moments accountant method when used as the privacy accountant method in DP-NF.  For each number of iterations, GDP accounting yields a lower $\varepsilon$ privacy value.

\subsection{DP-SGD}\label{app.dpsgd}

Algorithm \ref{alg:dp-sgd} is a formal description of DP-SGD of \cite{Abadi_2016}, from which our approach is based. Specifically, Algorithm \ref{alg:DP-NF} is an instantiation of DP-SGD, substituting negative log likelihood for the generic loss function in Algorithm \ref{alg:dp-sgd}.

\begin{algorithm}[H]
\caption{DP-SGD, differentially private stochastic gradient descent \cite{Abadi_2016}}
\label{alg:dp-sgd}
\begin{algorithmic}[1]
\STATE \textbf{Input:} Dataset $\bm{X} = \{\bm{x}^{(1)}, \ldots, \bm{x}^{(n)}\}$, loss function $\mathcal{L}(\theta)$, learning rate $\eta_t$, noise multiplier $\sigma$, batch size $L$, gradient norm bound $C$.
\STATE \textbf{Initialize} $\theta_0$ randomly
\FOR{$t \in [T]$}
    \STATE Take a Poisson random subsample $L_t$ with per-example probability $L / N$.
    \STATE \textbf{Compute gradient}
    \STATE for each $i \in L_t$, compute $\bm{g}_t(x_i) \leftarrow \nabla_{\theta_t} \mathcal{L}(\theta_t, x_i)$
    \STATE \textbf{Clip gradient}
    \STATE $\bm{\bar{g}}_t(x_i) \leftarrow \bm{\bar{g}}_t(x_i) / \max(1, ||\bm{g}_t(x_i)||_2 / C)$
    \STATE \textbf{Add noise}
    \STATE $\bm{\tilde{g}}_t(x_i) \leftarrow \frac{1}{L}(\sum_i \bm{\bar{g}}_t(x_i) + \mathcal{N}(\bm{0}, \sigma^2 C^2 \bm{I}))$
    \STATE \textbf{Descend}
    \STATE $\theta_{t+1} \leftarrow \theta_t - \eta_t \bm{\tilde{g}}_t(x_i)$
\ENDFOR
\STATE \textbf{Output} $\bm{\theta}_T$ and compute overall $(\varepsilon, \delta)$ using a privacy accounting method.
\end{algorithmic}
\end{algorithm}

\section{DP-NF Extensions}\label{app.extend}

\subsection{Data-Dependent Initialization of Normalization Layers}\label{s.norm}

Intermediate normalization layers such as batch normalization \cite{DBLP:journals/corr/IoffeS15} and activation normalization \cite{kingma2018glow} have been shown to improve the stability of normalizing flow models. In our context, batch normalization is incompatible with our approach since that batch statistics are shared when computing the forward pass of the layer, precluding the ability to calculate truly independent per-example gradients as required by NoisySGD. Activation normalization is more appropriate in our setting since no such batch statistics are calculated. Activation normalization is characterized by an offset and scaling of its inputs feature-wise by a learned set of parameters $\bm{b}$ and $\bm{w}$, i.e. $\bm{y}^{(i)} \gets (\bm{x}^{(i)} - \bm{b}) / \bm{w}$. In practice, typically these parameters are set via data-dependent initialization \cite{DBLP:journals/corr/SalimansK16} by computing a forward pass on a sampled batch of data and setting $\bm{b}$ and $\bm{w}$ to be the per-feature means and standard deviations of the inputs it had observed respectively. Since these statistics are calculated directly from the data, this approach is not privacy-preserving. 

One potential approach to making differentially private activation normalization is to privatize these statistics using the Laplace Mechanism \cite{DMNS06}.  This approach is outlined in Algorithm \ref{alg:DP-NF-INIT}, where $clip(\bm{X}, \tilde{c})$ clips the values of $\bm{X}$ to be in the range $\left[ -\tilde{c}/2, \tilde{c}/2 \right]$, $\mu(\bm{X})$ computes the feature-wise mean of $\bm{X}$, $\sigma(\bm{X})$ computes the feature-wise standard deviation of $\bm{X}$, and $R$ is some data-independent parameter initialization method which maps standardized inputs to standardized outputs in expectation, e.g., He initialization \cite{DBLP:journals/corr/HeZR015}.

\begin{algorithm}
\caption{DP-NF-INIT, data dependent initialization of activation normalization layers}
\label{alg:DP-NF-INIT}
\begin{algorithmic}[1]
\STATE \textbf{Input:} Dataset $\bm{X} = \{\bm{x}^{(1)}, \ldots, \bm{x}^{(n)}\}$, transformation $f$ (e.g. MADE \cite{DBLP:journals/corr/GermainGML15}), number of layers $K$, initialization privacy budget $\varepsilon$, initialization privacy tolerance $\delta$, data-independent parameter initialization method $R$ (e.g. He initialization \cite{DBLP:journals/corr/HeZR015}).
\STATE $\{\bm{\theta}_1, \ldots, \bm{\theta}_K\} \gets R()$
\FOR{$k = 1, \ldots, K$}
    \STATE $\bm{X} \gets clip(f_{\bm{\theta}^{(k)}}(\bm{X}), \tilde{c})$
    \STATE $\bm{b}^{(k)} \gets \mu(\bm{X}) + Lap(\frac{2\sqrt{4K \ln(1/\delta)} \triangle \hat{\mu}}{\varepsilon})$
    \STATE $\bm{w}^{(k)} \gets \sigma(\bm{X}) + Lap(\frac{2\sqrt{4K \ln(1/\delta)} \triangle \hat{\sigma}}{\varepsilon})$
    \STATE $\bm{X} \gets (\bm{X} - \bm{b}^{(k)}) / \bm{w}^{(k)}$
\ENDFOR
\STATE $\textbf{Output} \text{ concatenation of } \bm{\theta}^{(1)}, \bm{b}^{(1)}, \bm{w}^{(1)}, \ldots, \bm{\theta}^{(K)}, \bm{b}^{(K)}, \bm{w}^{(K)}$
\end{algorithmic}
\end{algorithm}

We note that Algorithm \ref{alg:DP-NF-INIT} is far from the only approach for differentially private activation normalization.  If the analyst has some domain knowledge about appropriate ranges of these parameters, she could use the differentially private Propose-Test-Release framework \cite{DL09} to first normalize $\bm{X}$ and then add noise proportionally to her proposed (and tested) sensitivity. While this approach seems practical for the outer layers, it is unlikely that an analyst would have numerical intuition for appropriate parameter values in all layers (especially when $K$ is large).  Thus even though the noise addition scheme described in Algorithm \ref{alg:DP-NF-INIT} may seem naive, it is likely the most practical approach.

Although the utility of activation normalization layers is quite evident, the original work \cite{kingma2018glow} proposing such layers provided little evidence to support the idea that data-dependent initialization yielded statistically significant improvements over a default initialization scheme, i.e., $\bm{b} \gets \bm{0}$ and $\bm{w} \gets \bm{1}$. In our experiments, we observed little distinction in contexts where the input data was assumed to be standardized and parameters were initialized to maintain variance between layers. Despite this, we include the approach for completeness for potential future contexts where data-dependent initialization of such parameters deems necessary.

\subsection{DP-MoG as a Prior}\label{s.dpem}

Thus far it has been assumed that we use the spherical multivariate Gaussian distribution to act as a prior for our model. Although, naturally any distribution could act as such a prior as long as it exhibits a tractable density function. For example, simply extending the single standardized Gaussian to a mixture of Gaussians has been shown \cite{NIPS2017_6828} to exhibit modest performance improvements. This mixture could be fit to the data \cite{izmailov2019semisupervised} a priori as well.

Hence, a natural extension to our proposed approach would be to fit DP-MoG first with privacy budget $(\varepsilon_1, \delta_1)$ to act as a prior, and then to refine this prior by training a sequence of nonlinear bijective functions with privacy budget $(\varepsilon_2, \delta_2)$ to yield an encompassing normalizing flow model. This yields a worst-case $(\varepsilon_1 + \varepsilon_2, \delta_1 + \delta_2)$-differential private guarantee by sequential composition. Although, this guarantee is easily improved by composing these privacy guarantee under some alternative privacy definition (e.g. RDP or GDP) before subsequently converting to a corresponding ($\varepsilon, \delta$)-DP guarantee. One might hypothesize that this approach would yield preferable results in contexts where the distribution at hand is composed of several discontinuous components, while exhibiting locally nonlinear density within each component.

\begin{figure}[htbp!]
  \centering
  \includegraphics[width=0.3\columnwidth]{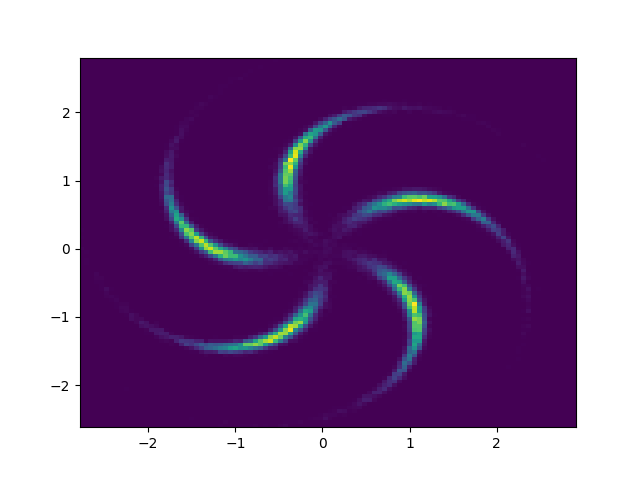}
  \caption{Pinwheel Dataset}
  \label{fig:pinwheel}
\end{figure}

To capture this context, we evaluate the efficacy of this approach on the Pinwheel dataset, as illustrated in Figure \ref{fig:pinwheel} (further details of this dataset are given in Appendix \ref{app.results}). The Pinwheel dataset is a common density estimation benchmark consisting of a number of disconnected components with nonlinear boundary. A Gaussian mixture model would naturally have difficulty approximating such a distribution for a small number of components, while classical normalizing flow models with a single standardized Gaussian as a prior might have difficulty expressing its discontinuous density.

As shown in Table \ref{pinwheellikelihoods}, using a trained GMM to act as a prior can aid in performance. First, we note that both DP-NF and DP-MoG demonstrate difficulty in achieving negative log likelihoods lower than 2.65-2.70, even when the number of components for DP-MoG is increased. DP-NF with DP-MoG as a prior modestly outperforms both alternatives used in isolation. Additionally, if one assumes that a GMM prior of five components fit to the population can be assumed to be public, one achieves dramatic performance improvements over all methods.

\begin{table*}[h]
  \caption{Test negative log likelihood (lower is better) on Pinwheel dataset (Figure \ref{fig:pinwheel}) for varying privacy budgets $\varepsilon$, composed via the moments accountant. \textbf{From top to bottom}: Standard DP-NF with a single spherical Gaussian as a prior, DP-NF with a GMM prior of 5 components (fit non-privately), DP-NF with a GMM prior of 5 components (learned privately for $\varepsilon = 0.2, \delta = 0$), DP-MoG with 5 components, DP-MoG with 10 components. When a privacy budget is expended on the prior, this was included when calculating the overall cost indicated by the column headings.}
  \centering
  \begin{tabular*}{0.7\textwidth}{lrrrrr}
    \toprule
        \textbf{Pinwheel} \\
        $\delta = 3.70 \times 10^{-5}$ & $\varepsilon = 1.50$ & $\varepsilon = 2.50$ &  $\varepsilon = 3.50$ &  $\varepsilon = 4.50$ \\
        \midrule
        DP-NF\textsubscript{spherical Gaussian prior}                                      & $2.80$ & $2.78$ & $2.74$ & $2.63$ \\
        DP-NF\textsubscript{non-private GMM prior}                                         & $3.05$ & $2.40$ & $1.91$ & $1.87$ \\
        DP-NF\textsubscript{private GMM prior, $\varepsilon=0.2$}                          & $2.77$ & $2.70$ & $2.62$ & $2.47$ \\
        DP-MoG\textsubscript{$5$}                                                           & $2.75$ & $2.76$ & $2.76$ & $2.76$ \\
        DP-MoG\textsubscript{$10$}                                                          & $2.76$ & $2.88$ & $2.77$ & $2.77$ \\
    \bottomrule
  \end{tabular*}
  \label{pinwheellikelihoods}
\end{table*}

\section{Additional Results and Training Details}\label{app.results}

In this section, we provide additional results on the performance of DP-NF, compared to the baseline mechanism of DP-MoG, as evaluated on several real and synthetic datasets.  These datasets are summarized in Table \ref{tab.datasets}.  Details of the real and synthetic datasets are respectively given in Sections \ref{app.real} and \ref{app.syn}.  Further details of hyperparameter training for all experiments are given in Section \ref{app.training}.

\begin{table}[htbp]
  \caption{Dimensionality $D$ and number of examples $n$ for each dataset.}
  \centering
  \begin{tabular}{lcrr}
    \toprule
        Dataset      & Real &  $D$ &       $n$ \\
    \midrule
        Life Science & Real &   10 &    26,733 \\
        Gowalla      & Real &    2 &   100,000 \\
        Power        & Real &    6 &   100,000 \\
        Gaussians    & Synthetic &    2 &    30,000 \\
        Half-Moons   & Synthetic &    2 &    30,000 \\
        Pinwheel     & Synthetic &    2 &    30,000 \\
    \bottomrule
  \end{tabular}
\label{tab.datasets}
\end{table}

\subsection{Additional Results on Real Datasets}\label{app.real}

Our analysis aims to cover a range of datasets composed of both synthetic and real-world datasets. In this section, we provide short descriptions of each dataset used for evaluation, and provide results of DP-NF evaluated on these datasets.

\textbf{Life Science}. The Life Science dataset is a density estimation benchmark dataset from the UCI machine learning repository \cite{Dua:2019} used in our baseline \cite{pmlr-v54-park17c} in their evaluation of DP-MoG. It contains 26,733 real-valued records of dimension 10 characterizing the principle components of measurements made in a variety of chemical and biological experiments.

\textbf{Power}. The Power dataset is a density estimation benchmark dataset from the UCI machine learning repository \cite{Dua:2019} used in much of the normalizing flow literature \cite{NIPS2017_6828, DBLP:journals/corr/abs-1810-01367}. It contains measurements of electric power consumption in a household over a period of 47 months, and was preprocessed according to the description given in \cite{NIPS2017_6828}.

\textbf{Gowalla}. The Gowalla dataset contains the locations in terms of longitude and latitude of the social network's users' check-ins. The total number of points is 1,256,384, which was reduced to 100,000 via a random sample. It was used in the evaluation of our baseline \cite{pmlr-v54-park17c}, but applied to the task of $k$-means clustering rather than learning the components of a Gaussian mixture model.

We evaluated the performance of DP-NF and the baseline DP-MoG for comparison on these real-world datasets.  Results on the Life Science dataset are given in the body of the paper in Table \ref{fig:likelihoods}.  Results on the Power and Gowalla datasets are given in Table \ref{real.likelihoods} below.

\begin{table}[htbp!]
  \caption{Average test log likelihood for varying privacy budgets $\varepsilon$. Error bars denote standard deviation over ten independent cross-validation splits. DP-MoG\textsubscript{$x$} refers to a Gaussian mixture of $x$ components. Omitted entries occur when model training was early-stopped because it had already converged at lower $\varepsilon$ values.}
  \centering
  \begin{tabular}{lrrrr}
    \toprule
        \textbf{Power} \\
        $\delta = 1.11 \times 10^{-5}$ & $\varepsilon = 1.25$ &  $\varepsilon = 1.75$ &  $\varepsilon = 2.25$ \\
        \midrule
        DP-NF (GDP)                    &   $-2.12 \pm 0.29$ &           -        &          -         \\
        DP-NF (MA)                     &   $-3.43 \pm 0.19$ &       $-2.47 \pm 0.13$ &       $-2.16 \pm 0.25$ \\
        \midrule
        DP-MoG\textsubscript{10} (MA)    &     $-2.88 \pm 0.04$ &       $-2.87 \pm 0.04$ &       $-2.86 \pm 0.04$ \\
        DP-MoG\textsubscript{10} (zCDP)  &     $-3.29 \pm 0.04$ &       $-3.10 \pm 0.04$ &       $-3.04 \pm 0.04$ \\
    \midrule
        \textbf{Gowalla} \\
        $\delta = 1.11 \times 10^{-5}$ &  $\varepsilon = 2.25$ &  $\varepsilon = 3.50$ &  $\varepsilon = 4.00$ \\
        \midrule
        DP-NF (GDP)                    &       $-0.44 \pm 0.35$ &               -    &          -         \\
        DP-NF (MA)                     &       $-1.81 \pm 0.11$ &       $-0.77 \pm 0.30$ &       $0.47 \pm 0.32$ \\
        \midrule
        DP-MoG\textsubscript{11} (MA)    &     $-0.92 \pm 0.01$ &       $-0.92 \pm 0.01$ &       $-0.91 \pm 0.01$ \\
        DP-MoG\textsubscript{11} (zCDP)  &     $-0.87 \pm 0.01$ &       $-0.93 \pm 0.01$ &       $-0.93 \pm 0.01$ \\
    \bottomrule
  \end{tabular}
  \label{real.likelihoods}
\end{table}

\begin{figure}[h]
    \centering
    \includegraphics[width=\linewidth]{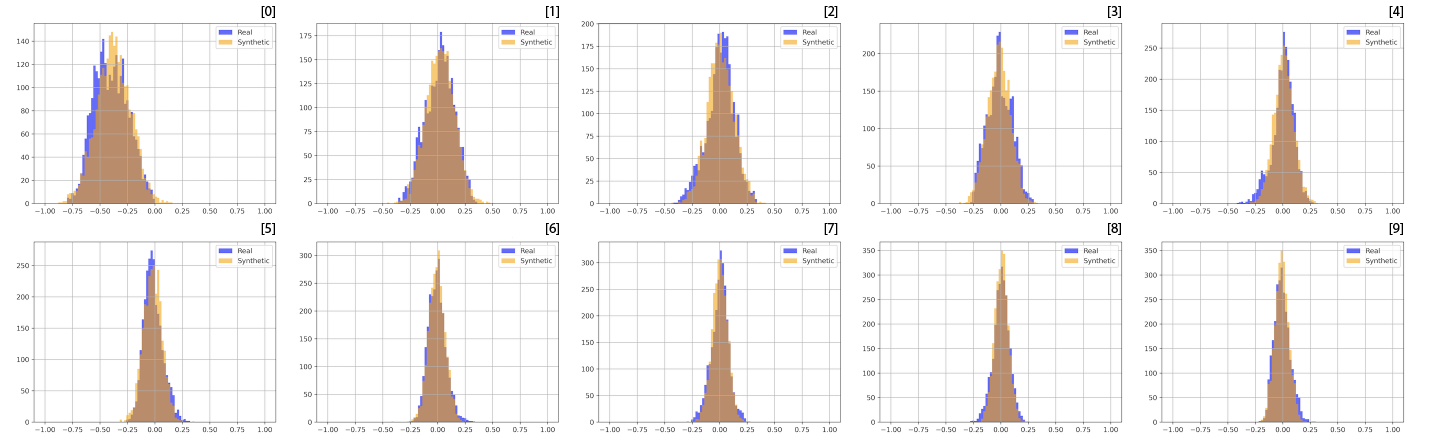}
    \includegraphics[width=\linewidth]{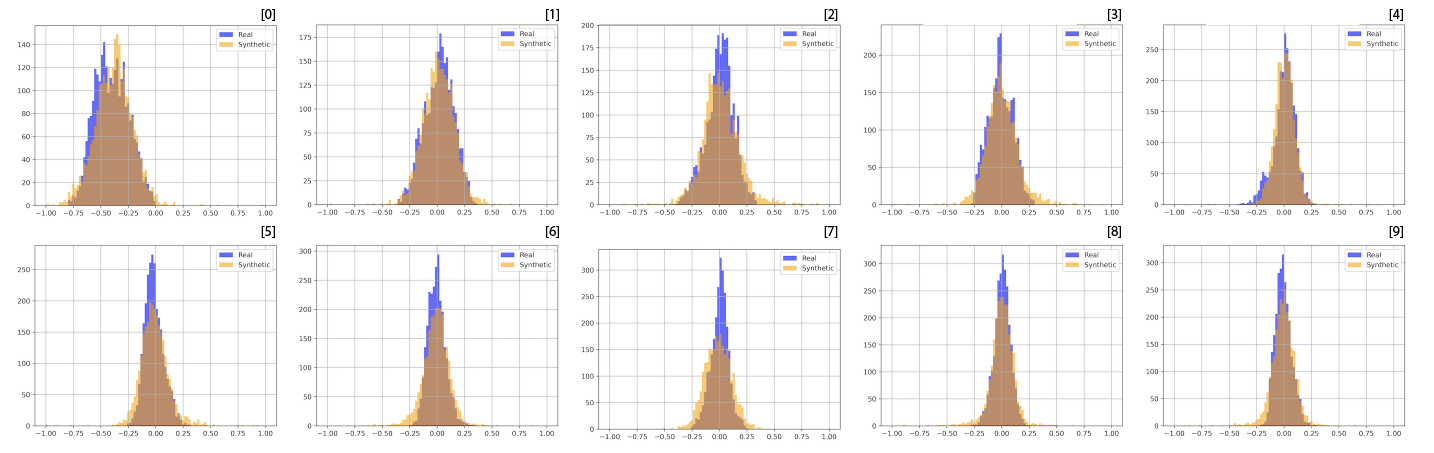}
    \caption{Dimension-wise histograms of synthetically generated Life Science data, superimposed over real data, for $\varepsilon = 0.5$ and $\delta = 1.52 \times 10^{-5}$. Dimensions are given in order from 0 to 9 left-to-right, top-to-bottom. \textbf{Top two rows:} DP-NF. \textbf{Bottom two rows:} DP-MoG. Note that synthetic data from DP-NF represents the real data well, while DP-MoG is relatively unable to to capture concentrated regions of density in the real data.}
    \label{fig:lifesci-samples-dimwise}
\end{figure}

We also provide Figure \ref{fig:lifesci-samples-dimwise}, which is the full version of Figure \ref{fig:lifesci-samples-dimwise-nf} in Section \ref{s.exp}. Figure \ref{fig:lifesci-samples-dimwise} provides dimension-wise histograms of the synthetically generated data for all ten dimensions of the Life Science dataset, presented in numerical order by axis index from left to right. The top two rows correspond to our baseline, and the bottom two rows correspond to our approach. For each image, we visualize the synthetically generated data (given in orange) and superimpose it over real data (given in blue) for comparison. We observe that for nearly all ten features, the distribution of data generated by DP-NF closely resembled that of the real data while DP-MoG was unable to replicate regions of concentrated density for certain dimensions.

\subsection{Additional Results on Synthetic Datasets}\label{app.syn}

\begin{figure}[H]
\centering
    \includegraphics[width=0.3\columnwidth]{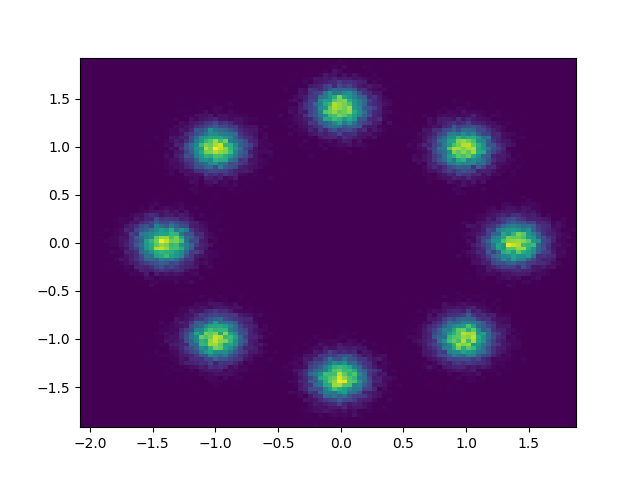}
    \caption{Gaussian Dataset}\label{fig:gaussian}
\end{figure}

We also evaluate our DP-NF method on several synthetic datasets. We perform this on the Half-Moons dataset (Figure \ref{fig:moons}), as well as a synthetically constructed dataset of a mixture of 8 Gaussians (Figure \ref{fig:gaussian}). This is done to demonstrate the heightened expressiveness of our approach as compared to a Gaussian mixture approach, alongside a worst-case scenario where the data is truly generated by a mixture of Gaussians, where DP-MoG would be expected to outperform our method. Results are presented in Table \ref{synth.likelihoods}.


\begin{table}[h!]
  \caption{Average test log likelihood for varying privacy budgets $\varepsilon$. Error bars denote standard deviation over ten independent cross-validation splits. DP-MoG\textsubscript{$x$} refers to a Gaussian mixture of $x$ components. Omitted entries occur when model training was early-stopped because it had already converged at lower $\varepsilon$ values.}
  \centering
  \begin{tabular}{lrrrrr}
    \toprule
      \textbf{Half-Moons} \\
      $\delta = 3.70 \times 10^{-5}$ & $\varepsilon = 1.50$ &  $\varepsilon = 2.25$ &  $\varepsilon = 3.00$ \\
      \midrule
      DP-NF (GDP)                    &      $-2.23 \pm 0.06$ &                  -     &                 -      \\
      DP-NF (MA)                     &       $-2.57 \pm 0.04$ &       $-2.43 \pm 0.06$ &       $-2.22 \pm 0.06$ \\
      \midrule
      DP-MoG\textsubscript{3} (MA)    &      $-2.59 \pm 0.01$ &       $-2.58 \pm 0.01$ &       $-2.58 \pm 0.01$ \\
      DP-MoG\textsubscript{3} (zCDP)  &      $-2.73 \pm 0.01$ &       $-2.60 \pm 0.01$ &       $-2.60 \pm 0.01$ \\
    \midrule
      \textbf{Gaussians} \\
      $\delta = 3.70 \times 10^{-5}$ & $\varepsilon = 1.50$ &  $\varepsilon = 2.25$ &  $\varepsilon = 3.00$ \\
      \midrule
      DP-NF (GDP)                    &      $-2.45 \pm 0.11$ &                  -     &                 -       \\
      DP-NF (MA)                     &      $-2.73 \pm 0.02$ &       $-2.61 \pm 0.08$ &        $-2.45 \pm 0.11$ \\
      \midrule 
      DP-MoG\textsubscript{8} (MA)    &      $-2.44 \pm 0.01$ &       $-2.45 \pm 0.01$ &        $-2.45 \pm 0.01$ \\
      DP-MoG\textsubscript{8} (zCDP)  &      $-2.42 \pm 0.01$ &       $-2.41 \pm 0.01$ &        $-2.43 \pm 0.01$ \\
    \bottomrule
  \end{tabular}
  \label{synth.likelihoods}
\end{table}

\subsection{Hyperparameters and Training details}\label{app.training}

In this subsection we detail decisions about hyperparameter section in training, including the gradient clipping parameter $C$, regularization of the loss function, and choice of privacy accountants.

For the gradient clipping parameter $C$, prior work \cite{Abadi_2016} suggested that a reasonable heuristic for setting this clipping parameter was to set $C$ equal to the median of the $\ell_2$ norms of the unclipped gradients observed over the course of a non-private training execution. In the context of normalizing flows, we found that much larger values for $C$ yielded significantly preferable results. A natural explanation for this arises under consideration of log likelihood as an objective. In cases where a given point is assigned near-zero density, a large gradient update would be incurred to prevent further deterioration. When this gradient update is clipped, the resulting update may be insufficient to avoid associated numerical instability if this point is assigned density further approaching zero. Despite excess noise being applied to updates with larger $C$, we found this to merely prolong training without a significant degradation in resulting model quality.

With respect to regularization, \cite{NIPS2017_6828} suggested a modest amount of $\ell_2$ regularization (i.e., a coefficient of $10^{-6}$) in the context of non-private normalizing flows. We found this regularization approach to substantially degrade the quality of the resulting models and was generally omitted in our training. This makes intuitive sense, as the suggested regularization of \cite{NIPS2017_6828} serves to decrease model weights over the course of training, and differentially private optimization applies Gaussian noise vectors of constant variance to gradients throughout training. As model weights tend toward zero, one would expect that the noise injection from privacy eventually dominates the learned criteria of the model.


For privacy accountants, we found that composition under Gaussian differential privacy (GDP) \cite{DBLP:journals/corr/abs-1905-02383} consistently yielded the tightest privacy bounds throughout our experiments. See Figure \ref{fig:lifesci-privacy-guarantees} in Appendix \ref{app.gdp} for an illustration of the improvements in privacy composition that can be achieved by GDP, over the moments accountant of \cite{Abadi_2016}. Since our baseline method DP-MoG yielded the best performance under the moments accountant in \cite{pmlr-v54-park17c}, we included both GDP composition and moments accountant for fair comparison. We found that DP-NF consistently outperformed DP-MoG even when both methods used the moments accountant, and that further performance improvements could be achieved by DP-NF using GDP composition as a privacy accountant method.


\section{Limitations of Proposed Approach}

Normalizing flow models are trained in a manner which minimizes the average negative log likelihood of the observed data. As such, it is not uncommon when training such models that a given point is assigned near-zero density, provoking a loss approaching infinity. We observed this issue was somewhat exacerbated due to differentially private optimization, in part due to noise injection, but primarily as a result of subsampling. In the case of uniform subsampling, as required in our privacy analysis, there is no strict guarantee that any individual point is regularly sampled. This acts in contrast to typical sampling methodology in which the dataset is repeatedly shuffled and partitioned into equal sized batches over the course of an epoch. Rigorous privacy guarantees associated with equally sized and disjoint sampling is, to the best of our knowledge, a currently unsettled issue \cite{mcmahan2018general} and a potential avenue for improvement given its theoretical convergence guarantees. We did not find this limitation to be ultimately confounding in any way.

Additionally, given that DP-NF is ultimately a deep learning based approach to density estimation, it naturally involves the optimization of a large number of parameters and a high resource expenditure in terms of time and space complexity. This is especially highlighted in comparison to DP-MoG as a baseline, which takes on the order of seconds to run on CPU as compared to our method which can take on the order of an hour on GPU. Although, we find this tradeoff between resource expenditure and distribution quality to be justified, particularly in the context of differentially private data analysis and social science applications which rarely demand strict resource constraints within reason.

\end{document}